\documentclass[10pt,twocolumn,letterpaper]{article}

\usepackage[final]{cvpr}              

\definecolor{cvprblue}{rgb}{0.21,0.49,0.74}
\usepackage{adjustbox}
\usepackage[pagebackref,breaklinks,colorlinks,allcolors=cvprblue]{hyperref}
\usepackage{multirow}

\definecolor{teal}{rgb}{0, 0.7, 0.6}

\def\confName{CVPR}
\def\confYear{2025}

\title{Multitwine: Multi-Object Compositing with Text and Layout Control}  

\author{Gemma Canet Tarrés \\ University of Surrey \\ \tt\small g.canettarres@surrey.ac.uk \and Zhe Lin \\ Adobe Research \\ \tt\small zlin@adobe.com \and Zhifei Zhang \\ Adobe Research \\ \tt\small zzhang@adobe.com \and He Zhang \\ Adobe Research \\ \tt\small hezhan@adobe.com \and Andrew Gilbert \\ University of Surrey \\ \tt\small a.gilbert@surrey.ac.uk \and John Collomosse \\ University of Surrey \& Adobe Research \\ \tt\small j.collomosse@surrey.ac.uk \and Soo Ye Kim \thanks{corresponding authors. This work was done 50/50 in collaboration between Adobe and University of Surrey. Partially supported by DECaDE under EPSRC Grant EP/T022485/1.} \\ Adobe Research \\ \tt\small sooyek@adobe.com}

\author{Gemma Canet Tarrés$^{1}$\\
{\tt\small g.canettarres@surrey.ac.uk}
\and
Zhe Lin$^2$\\
{\tt\small zlin@adobe.com}
\and
Zhifei Zhang$^2$\\
{\tt\small zzhang@adobe.com}
\and
He Zhang$^2$\\
{\tt\small hezhan@adobe.com}
\and
Andrew Gilbert$^{1*}$\\
{\tt\small a.gilbert@surrey.ac.uk}
\and
John Collomosse$^{1,2*}$\\
{\tt\small j.collomosse@surrey.ac.uk}
\and
Soo Ye Kim$^{2*}$\\
{\tt\small sooyek@adobe.com}
\and
$^1$University of Surrey \qquad $^2$Adobe Research
}

\begin{document}


\twocolumn[{%
\renewcommand\twocolumn[1][]{#1}%
\maketitle
\vspace{-15pt} 
    \includegraphics[width=\linewidth]{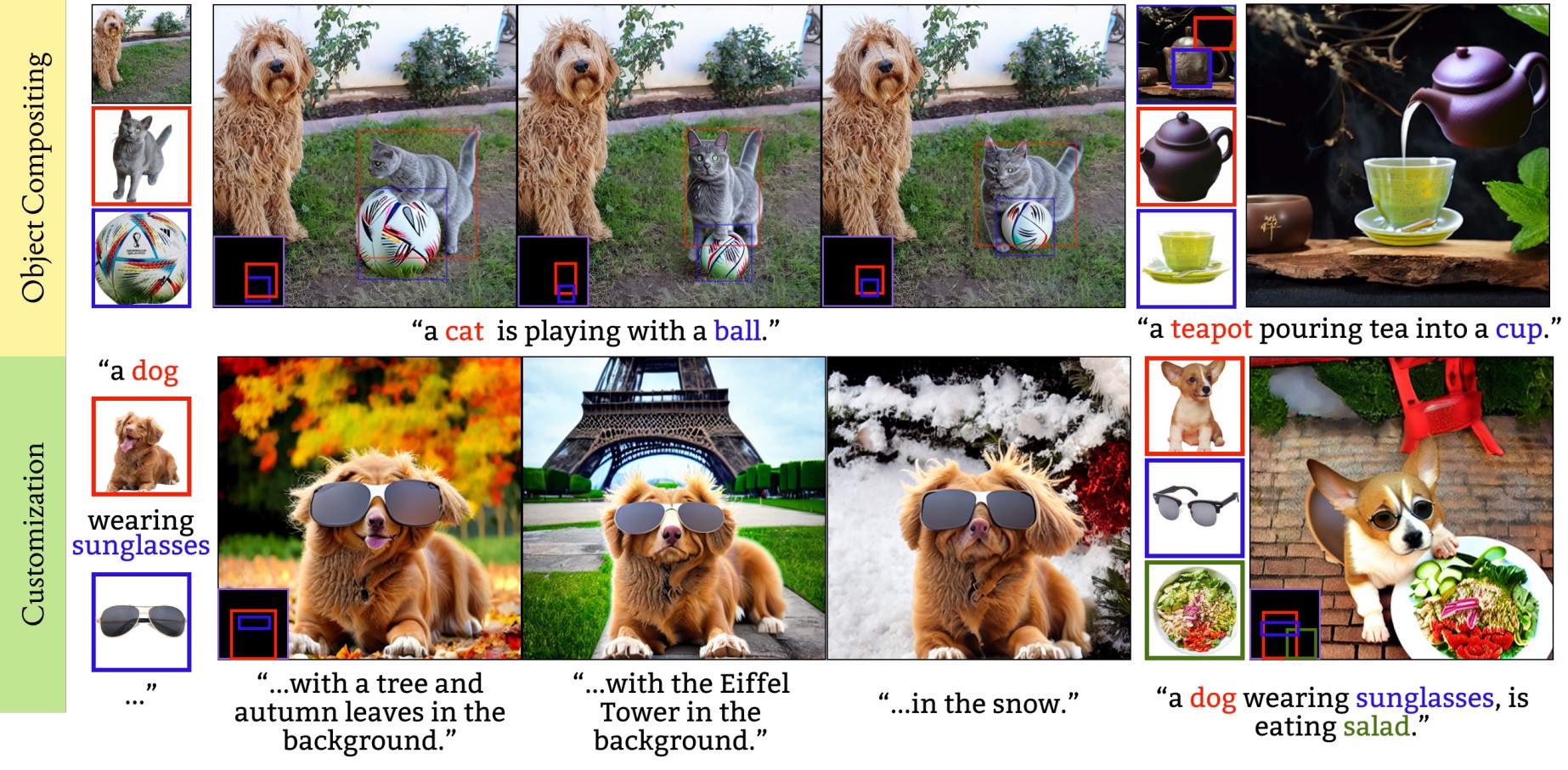}
    Figure 1. Our model allows simultaneous compositing of several objects with text and layout control, offering harmonious and natural results (\textit{top row}). Object relations and actions can be controlled through text, allowing reposing and automatic addition of necessary objects (\ie flowing tea). Pose and layout can be controlled via bounding boxes provided as input. Our model can also be used for customization (\textit{bottom row}), seamlessly integrating given objects into described backgrounds.
    \vspace{7.5pt}
    \label{fig:teaser}
}]

\setcounter{figure}{1}

\renewcommand{\thefootnote}{\fnsymbol{footnote}}
\footnotetext[1]{: corresponding authors. This work was a joint collaboration between Adobe and the University of Surrey, conducted during an internship of the main author at Adobe. It was partially supported by DECaDE under EPSRC Grant EP/T022485/1.} 

\begin{abstract}

We introduce the first generative model capable of simultaneous multi-object compositing, guided by both text and layout. Our model allows for the addition of multiple objects within a scene, capturing a range of interactions from  simple positional relations (\eg, \textit{next to, in front of}) to complex actions requiring reposing (\eg, \textit{hugging, playing guitar}). When an interaction implies additional props, like `taking a selfie', our model autonomously generates these supporting objects. By jointly training for compositing and subject-driven generation, also known as \textit{customization}, we achieve a more balanced integration of textual and visual inputs for text-driven object compositing. As a result, we obtain a versatile model with state-of-the-art performance in both tasks. We further present a data generation pipeline leveraging visual and language models to effortlessly synthesize multimodal, aligned training data.

\end{abstract}    
\section{Introduction}
\label{sec:intro}

\begin{figure}[t]
    \centering
    \includegraphics[width=\linewidth]{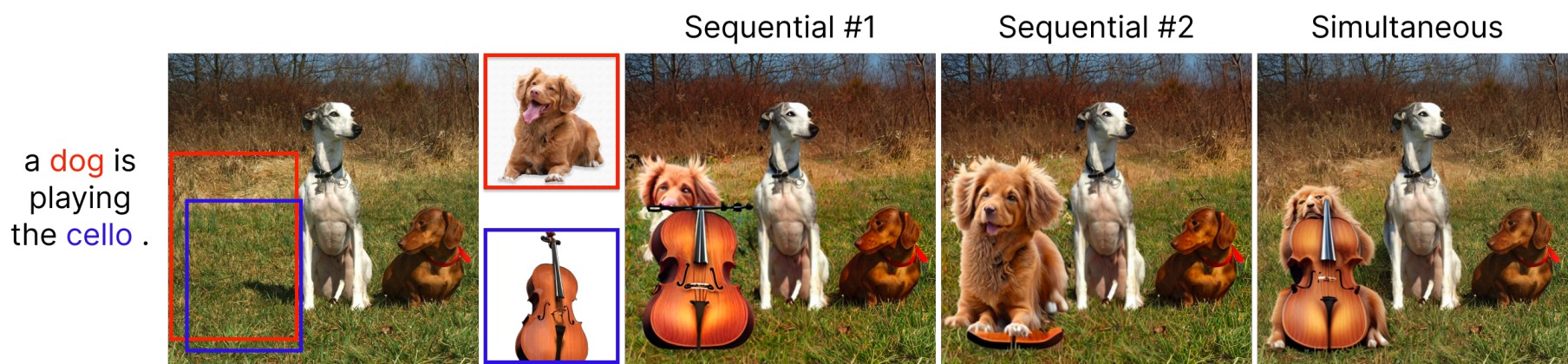}
    \caption{Comparison of simultaneous vs. sequential object compositing. Sequential addition prevents reposing of previously composited objects, resulting in limited, less cohesive compositions.} 

    \label{fig:motivation}
    \vspace{-6mm}
\end{figure}

In recent years, advancements in image generation and editing have enabled the creation of realistic and complex visual scenes from various input modalities, such as text descriptions \cite{rombach2022ldm,podell2023sdxl,esser2024scaling}. These tools are becoming essential for enhancing creativity and streamlining workflows. A key component in this process is object compositing, enabling seamless integration of new objects in an existing scene. However, current compositing models \cite{song2024imprint,tarres2024thinking,yang2023paintbyexample,zhang2023controlcom,chen2023anydoor} are generally limited to handling a single object at a time, requiring sequential steps to composite multiple objects into a scene. This process is cumbersome and fails to capture the complex, real-world interactions between multiple objects, especially when simultaneous repositioning is necessary. For instance, sequential compositing struggles to accurately create interactions where objects need to interact very closely, such as two figures hugging, or someone playing an instrument (Fig \ref{fig:motivation}). 

To address these limitations , we introduce a novel text-guided and layout-guided multi-object compositing model designed to handle and harmonize multiple objects simultaneously within a single composition. This model bridges the gap between single object insertion and realistic, context-aware scene construction via three main features: (i) it allows for simultaneous reposing and complex interaction between objects, seamlessly capturing intricate relationships; (ii) it ensures visual coherence between both the new objects and the scene, achieved through synchronized relighting and reharmonization;  (iii) it automatically generates additional elements essential to the action or interaction being depicted (\eg, a leash for a dog-walking scenario, or liquid pouring from a bottle to a glass), creating a more cohesive and natural scene.

Our model accepts multimodal input, including object images, text, object-specific bounding boxes, a background image and a mask delimiting the overall compositing region. To enable effective and balanced integration of these modalities, our model is trained on diverse, multimodal data that provides rich grounding for scene context and object relationships, allowing it to understand and implement nuanced multi-object compositions. Collecting data with all these aligned modalities is a complex task. However, advancements in Large Language Models (LLMs) \cite{liu2023improvedllava,driess2023palm,alayrac2022flamingo} and Visual Language Models (VLMs) \cite{liu2023groundingdino} offer a solution. We introduce a data generation pipeline that leverages these models to synthesize essential missing data, enabling the creation of a fully aligned, multimodal training set.
By leveraging varied data sources during training and balancing real and synthetic data, our model learns to anticipate and position additional supporting objects, if needed, to ensure natural compositions. Furthermore, we jointly train the model on compositing and customization tasks to improve compositing performance and offer more versatility. This dual approach allows the model to separately focus on two key subtasks in multi-object compositing: (i) generating realistic interactions with strong text-image alignment, and (ii) improving inpainting, harmonization and relighting. 

Our contributions are as follows:

\begin{itemize}
    \item We present the first generative model for simultaneous multi-object compositing, addressing the limitations of sequential object compositing.
    \item We propose a novel data generation pipeline that combines real labeled data, synthetic captions from vision-language models, and grounding methods to align global and local captions with images for improved training.
    \item We leverage customization as 
    an auxiliary
    task to improve text and image alignment in compositing, resulting in a model capable of performing both tasks with performance comparable to state-of-the-art customization and generative object compositing models.
\end{itemize}

\section{Related Work}
\label{sec:SoA}


\textbf{Generative Object Compositing.} Generative object compositing has evolved from traditional methods relying on hand-crafted features \cite{lalonde2007photo}, 3D modeling \cite{kholgade20143d}, and rendering \cite{karsch2011rendering} to leveraging the efficiency of diffusion models. Modern diffusion-based approaches, such as ObjectStitch \cite{song2022objectstitch} and Paint by Example \cite{yang2023paintbyexample}, integrate object and background but struggle with identity preservation due to reliance on CLIP \cite{radford2021clip}. TF-ICON \cite{lu2023tficon} improves on identity retention with noise modeling and composite self-attention injection, though it lacks flexibility in object reposing. \cite{seyfioglu2024diffusechoose} and \cite{kulal2023putting} improve performance by either using a secondary U-Net encoder or focusing on human generation. Recent works like AnyDoor \cite{chen2023anydoor} and IMPRINT \cite{song2024imprint} use DINO v2 \cite{oquab2023dinov2} for stronger identity fidelity and support more flexible shape and pose control. ControlCom \cite{zhang2023controlcom} enables separate control over blending, harmonization, reposing and compositing, while CustomNet \cite{yuan2023customnet} allows control of viewpoint and location. \cite{tarres2024thinking} expands generation across the entire image, offering an unconstrained approach that simplifies sequential object additions. However, none of these methods support simultaneous multi-object compositing with interdependent reposing or text-based inputs for refined control, both essential for interactive, complex scenes.

\noindent
\textbf{Subject-Driven Generation} Building on recent advances in text-to-image generation, subject-driven approaches aim to customize images by integrating subject-specific visuals within text prompts. Approaches like DreamBooth \cite{ruiz2023dreambooth} and Textual Inversion \cite{gal2022inversion} fine-tune a model to recontextualize a specific subject based on text, while others \cite{chen2024suti,shi2023instantbooth,jia2023taming} bypass fine-tuning via large-scale upstream training. Re-Imagen \cite{chen2022re}, SuTI \cite{chen2024suti}, FastComposer \cite{xiao2024fastcomposer}, ELITE \cite{wei2023elite} provide image features from image encoders directly to the U-Net, as with text control. BLIP-Diffusion \cite{li2024blip} enables zero-shot generation by embedding objects into random backgrounds but remains limited in handling multiple entities. GroundingBooth \cite{xiong2024groundingbooth} provides a model for grounding text-to-image generation, specifying a bounding box for each object as part of the input. Leveraging Multimodal Large Language Models (MLLMs) \cite{wang2023cogvlm,driess2023palm,alayrac2022flamingo} offers efficient integration of text and visual inputs. GILL \cite{koh2024gill}, Emu \cite{sun2023emu}, and DreamLLM \cite{dong2023dreamllm} align MLLM outputs with diffusion encoders for combined visual-language generation, though struggle to retain fine-grained visual details due to alignment at semantic level. KOSMOS-G \cite{pan2023kosmos} and UNIMO-G \cite{li2024unimo} combine a MLLM as text and image encoder with a U-Net from Stable Diffusion as image decoder by either training a subnetwork that allows their integration or performing end-to-end training of the U-Net, respectively. Emu2Gen \cite{sun2024emugen} further adds support for bounding box guidance for each subject, enriching image layout control.

\section{Methodology}
\label{sec:methodology}

We propose a diffusion-based model capable of simultaneously compositing multiple objects into a background in a controlled way, guided by both a layout and a text prompt. Training this model effectively requires extensive paired data, including: (i) ground truth images containing multiple objects, (ii) text descriptions with grounding information for these objects, (iii) segmented images of the objects, and (iv) their bounding boxes in the ground truth image. For simplicity, we focus on two objects at a time, which may interact through complex actions or positional relationships in the ground truth image. We present model architecture  in Section \ref{sec:model}, training and inference strategies in Sections \ref{sec:training} and \ref{sec:inference}, and data generation pipeline in Section \ref{sec:datageneration}.  

\subsection{Model Architecture}
\label{sec:model}

\begin{figure}[t]
    \centering
    \includegraphics[width=\linewidth]{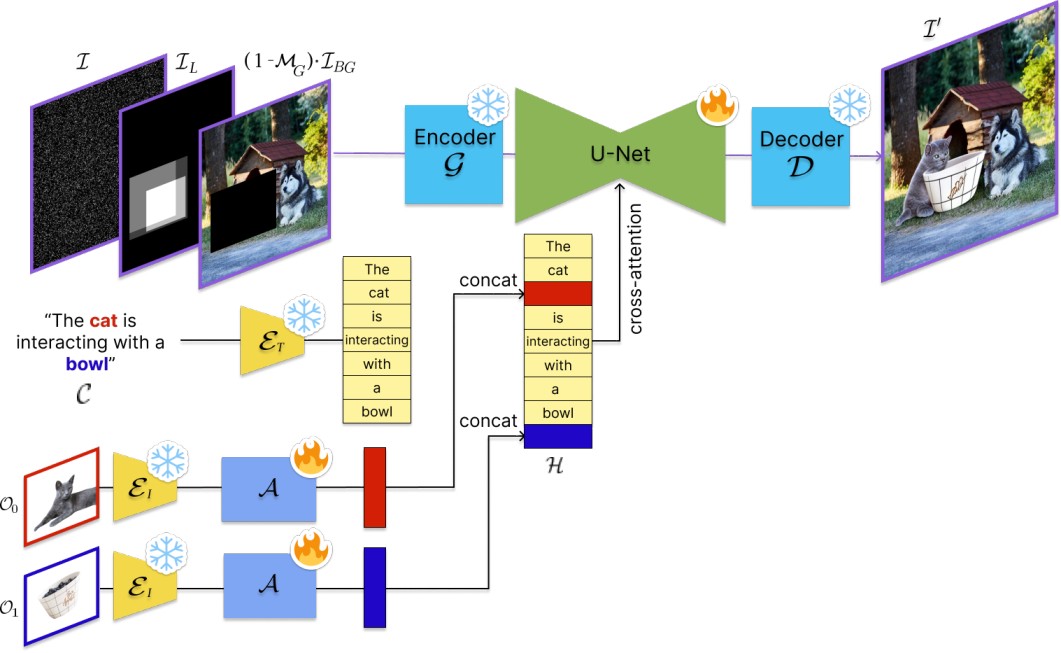}
    \caption{Model Architecture. Our model consists of: (i) A Stable Diffusion backbone including a U-Net and an autoencoder ($\mathcal{G}$, $\mathcal{D}$); (ii) a text encoder $\mathcal{E}_{T}$; (iii) an image encoder $\mathcal{E}_{I}$; and (iv) an adaptor $\mathcal{A}$. Given a text prompt $\mathcal{C}$ and images of N objects $\mathcal{O}_{0\dots, N-1}$, the text embedding from (iii) is augmented by concatenating each image embedding after their corresponding text tokens. The resulting multimodal embedding $\mathcal{H}$ is fed to the U-Net via cross-attention. Masked background image $(1-\mathcal{M}_{G})*\mathcal{I}_{BG}$ and layout $\mathcal{I}_{L}$ with object-specific bboxes are concatenated to input $\mathcal{I}$. }
    \label{fig:pipeline}
    \vspace{-6mm}
\end{figure}

Our model takes as input a background image $\mathcal{I}_{BG}$, a layout $\mathcal{I}_{L}$, $N$ object images $\mathcal{O}_{i}$, with $i \in \{0 \dots, N-1\}$, and a descriptive caption $\mathcal{C}$. The layout includes a bounding box for each object ($\mathcal{M}_{0 \dots, N-1}$), along with a larger box $\mathcal{M}_{G}$ that encloses them all, defining  the region to be modified (inpainting region), allowing for additional room for object-object interactions. 
We encode all layout information on a single mask, where different values are assigned to the pixels belonging to each object's bounding box, their overlapping, the rest of the inpainting region and the background. As shown in Fig \ref{fig:pipeline}, $\mathcal{I}_{L}$ is concatenated to the 3-channel input noise $\mathcal{I}$ and $(1-\mathcal{M}_{G})*\mathcal{I}_{BG}$, a version of $\mathcal{I}_{BG}$, where the inpainting region is masked out. These concatenated images are fed into the model's backbone, Stable Diffusion 1.5 (SD) \cite{rombach2022ldm}, consisting of a variational autoencoder ($\mathcal{G}$, $\mathcal{D}$) and a U-Net. Each composited object $\mathcal{O}_{i}$ is processed by an image encoder $\mathcal{E}_{I}$ (DINO ViT-G/14 \cite{oquab2023dinov2}) and a content adaptor $\mathcal{A}$, which aligns the object embeddings with text embeddings as in \cite{song2022objectstitch}. The caption $\mathcal{C}$ is encoded using a text encoder $\mathcal{E}_{T}$ (CLIP ViT-L/14 \cite{radford2021clip}). These embeddings are then combined into a multimodal embedding and fed into the U-Net via cross-attention.

\noindent
\textbf{Multimodal Embeddings} Providing the model with text and image information in a balanced and interpretable way is one of the main challenges of our method. It must perform well when either input is missing, and neither component must dominate when both are present. The identity of each object $\mathcal{O}_{i}$ must be preserved in the output while matching the scene and interactions described by the caption $\mathcal{C}$. Grounding information specifies the subset of words $\mathcal{C}_{i}$ in the caption $\mathcal{C}$ that correspond to each object image $\mathcal{O}_{i}$. After encoding the text, each object embedding $\mathcal{A}(\mathcal{E}_{I}(\mathcal{O}_{i}))$ is concatenated after $\mathcal{E}_{T}(\mathcal{C}_{i})$, resulting in a multimodal embedding $\mathcal{H}$ that is then passed to the U-Net via cross-attention. We refer to the set of visual and textual information referring to $\mathcal{C}_i$ and $\mathcal{O}_i$ in the embedding $\mathcal{H}$ as $\mathcal{H}_i$.



\subsection{Training Strategy}
\label{sec:training}

Our training strategy is designed to balance textual and visual information, enabling object reposing and the generation of described backgrounds and new objects while preserving each object's identity and preventing identity mixing. To ensure the model is still able to perform with only text or image input, we randomly drop each modality with a 30\% probability during training. 

\noindent
\textbf{Customization as an Auxiliary Task}  To enhance compositing performance and provide a more versatile model, we jointly train on multi-object compositing and multi-entity subject-guided generation. With 50\% probability, the inpainting region $\mathcal{M}_{G}$ is replaced by a mask covering the entire image, and $\mathcal{I}_{BG}$ is replaced with an empty image. In this customization setting, the model can focus on preserving object identity while reposing objects and generating scenes aligned with the text, without being burdened by compositing tasks like harmonization, relighting, or background inpainting. Introducing this joint training results in a better balance between textual and visual alignment. 

\noindent
\textbf{Object Identity Disentanglement} Diffusion models often struggle to accurately represent multiple subjects due to attention layers blending visual features and causing semantic leakage between them. Cross-attention maps indicate how text tokens influence latent pixels \cite{hertz2022prompt}. When tokens share similar semantics, a single pixel may respond to all, leading to identity blending. 
Additionally, self-attention features create dense correspondences within the same subject and across semantically similar ones \cite{dahary2024yourself,alaluf2024cross,cao2023masactrl}. While this behavior aids in generating coherent images with well-integrated subjects and backgrounds, it can also cause visual feature leakage between similar subjects \cite{dahary2024yourself}. To reduce this leakage and encourage identity separation, we introduce two additional losses ($\mathcal{L}_{c}$, $\mathcal{L}_{s}$) inspired by \cite{dahary2024yourself,xiao2024fastcomposer}:

\begin{equation}
\label{eq:crossattn-loss}
\mathcal{L}_{c} =  \sum\limits_{i = 0}^{N-1}\frac{1}{N}
\bigg(1 - \frac{
\sum\limits_{\substack{\mathbf{x} \in \mathcal{S}_i,\\ \mathbf{h} \in \mathcal{H}_i}}
\hat{\mathbf{A}}\left[ \mathbf{x}, \mathbf{h} \right]
}{
\sum\limits_{\substack{\mathbf{x} \in \mathcal{S}_i, \\ \mathbf{h} \in \mathcal{H}_i}}
\hat{\mathbf{A}}\left[ \mathbf{x}, \mathbf{h} \right]
+
\sum\limits_{\substack{\mathbf{x} \notin \mathcal{S}_i,\\ \mathbf{h} \in \mathcal{H}_i}}
\hat{\mathbf{A}}\left[ \mathbf{x}, \mathbf{h} \right]}\bigg)
,
\end{equation}


\begin{equation}
\label{eq:selfattn-loss}
\mathcal{L}_{s} =  \sum\limits_{i=0}^{N-1}\frac{1}{N}
\bigg(1 - \frac{
1}{1+\sum\limits_{\substack{\mathbf{x} \in \mathcal{S}_i, \\ \mathbf{y} \in \mathcal{S}_j, j \neq i}}
\hat{\mathbf{A}}\left[ \mathbf{x}, \mathbf{y} \right]
}\bigg)
,
\end{equation}

where $\hat{\mathbf{A}}$ is the mean attention map, averaged across heads and layers, $\mathbf{S}_i$ is the segmentation map of $\mathcal{O}_{i}$ in the ground truth image, and $\mathbf{x}$, $\mathbf{y}$ correspond to pixel coordinates in $\mathcal{I}_{t}$ (the noisy version of $\mathcal{I}$ at timestep $t$). For each  $\mathbf{x}$, $\mathcal{L}_{c}$ encourages the cross-attention maps obtained from visual-language information from $\mathcal{C}_i$ and $\mathcal{O}_i$ ($\mathbf{h} \in \mathcal{H}_i$) to be close to their corresponding segmentation map $\mathbf{S}_i$. $\mathcal{L}_{s}$ discourages pixels $\mathbf{x} \in \mathcal{S}_{i}$ to attend to pixels  $\mathbf{y} \in \mathcal{S}_{j}$, $\forall j \neq i$.

Based on the inpainting version of Stable Diffusion v1.5 \cite{rombach2022ldm}, our model is fine-tuned by optimizing a combination of three losses: $\mathcal{L} = \mathcal{L}_{d} + \alpha \mathcal{L}_{c} + \beta \mathcal{L}_{s}$. 

\begin{equation}
    \mathcal{L}_{d} = \mathbb{E}_{\mathcal{I}_{c}, \mathcal{H}, t,  \epsilon \sim \mathcal{N} (0,1)} \left[ \left\| \epsilon - \epsilon_\theta\left(\mathcal{I}_{c}, \mathcal{H}, t\right) \right\|_2^2 \right],
\label{eq:lossunet}
\end{equation}

where $\mathcal{I}_{c}=[\mathcal{I}_{t}, \mathcal{I}_{L}, (1-\mathcal{M}_{G})*\mathcal{I}_{BG}]$ ($\mathcal{I}_{t}$: noisy version of the input image $\mathcal{I}$ at timestep $t$, $\mathcal{I}_{L}$: layout, $(1-\mathcal{M}_{G})*\mathcal{I}_{BG}$: masked background image, $[\cdot]$: concatenation operation across the channel dimension), $\mathcal{H}$: multimodal embedding, $\epsilon_\theta$:denoising model being optimized.


\subsection{Inference Strategy}
\label{sec:inference}

Although attention-based losses help disentangle visual and semantic features of different objects during training, some information leakage can still occur at inference. To address this, and to further encourage the desired layout $\mathcal{I}_L$, all cross-attention scores corresponding to $\mathbf{h} \in \mathcal{H}_i$  are masked with $\mathcal{M}_i$. Additionally, since $[EoT]$ tokens are known to contain information about foreground objects \cite{zhao2023loco}, we mask their corresponding cross-attention scores with the union of all object masks $\bigcup_{i=0\dots N-1} \mathcal{M}_i$. 


\subsection{Training Data Generation}
\label{sec:datageneration}

Balanced, curated data is essential for effective diffusion model training. For our model to learn correct object interaction, natural compositing, text-image alignment and identity preservation, paired training data must meet specific criteria: (i) diverse range of objects and relationships; (ii) varied text prompt styles; (iii) object images with different poses, views and lighting than the ground truth image; (iv) high-quality images; and (v) a large dataset size. Since no single source can provide all these qualities, we use three complementary data sources: Video Data, In-the-Wild Images, and Manually Collected Data. 

\textbf{Video Data} We use annotated data from video object relation datasets \cite{shang2019annotating,shang2017video}, where each frame includes bounding boxes identifying objects and actions or positional relationships between object pairs in the format \textit{$<$object A$>$ $<$relation$>$ $<$object B$>$}. For each annotated relationship, we select one frame depicting it as the ground truth image, a second frame including \textit{object A} and a third with \textit{object B}. This provides alternative views for each object, which is essential for training the model to learn object reposing for intended interactions. To ensure identity consistency, we keep DINO score \cite{oquab2023dinov2} similarity for each object’s views above a threshold (MSE $\geq$ 0.8). We provide LLaVA \cite{liu2023improvedllava} with the ground truth image and the annotated relation and obtain a grammatically correct caption, used as aligned text prompt. Although this data source meets most of our training criteria, video data typically involves lower-quality frames.

\textbf{Image Data} An automated pipeline for paired data generation from in-the-wild images can provide unlimited data with high-quality images to enhance model training. We consider two approaches: top-down and bottom-up. \textbf{(i) Top-down approach:} Starting with a single image, we use a commercial subject selection tool to identify the main objects. A semantic segmentator \cite{qi2022entityseg} helps us select images where multiple key objects are present. From these, we randomly pick two entities to serve as composited objects. We outline each object in a different color on the original image and input it into ViP-LLaVA \cite{cai2024vip}, which identifies the selected entities and generates a caption describing their relationship. \textbf{(ii) Bottom-up approach:} To increase caption diversity, we include open-source images with ground-truth captions, such as those in OpenImages \cite{krylov2021open}. Given an image and its caption, GroundingDINO \cite{liu2023groundingdino} extracts grounding information that correlates the two. Using a semantic segmentator \cite{qi2022entityseg} and filtering for quality (e.g., removing duplicates, background entities, and objects that are too large ($>$75\%) or too small ($<$10\%)), we randomly select two grounded entities as composited objects. This second approach adds diversity to the image layouts and caption formats, which enhances model robustness through a mix of local and global descriptions. However, neither of the approaches provides different views of the objects, making them unsuitable as stand-alone sources for model training.

\textbf{Manually Collected Data} We manually compile a dataset containing 16,896 images of humans interacting with objects, 3,898 images of the same humans in neutral poses, and 1,250 images of those objects alone, captured from different angles and scenes. This data is segmented using \cite{qi2022entityseg} and their caption and grounding information is obtained using ViP-LLaVA \cite{cai2024vip}, following the same process described above. This collection provides a small curated set of high-quality images, including different views of each entity and covering human-object interactions, one of the most intricate relationships we aim to replicate.



\section{Experiments}
\label{sec:experiments}

\textbf{Evaluation Dataset} We evaluate image and text alignment in the customization task using the set of 300 paired two entities and text from \textit{MultiBench} \cite{li2024unimo}, a benchmark crafted for evaluating multi-entity subject-driven generation. We manually craft a bounding box for each category (\eg food, toy) to be used as input for our model and \cite{sun2024emugen}. For evaluating multi-object compositing, we create a test set of 118 paired data (background image, text prompt, two entities and their corresponding bounding boxes). We ensure different positional and action-based prompts are included, and entities include objects, animals and humans from \cite{pixabay,ruiz2023dreambooth,li2024unimo}. We also evaluate identity preservation on \textit{Dreambooth}, the single-object compositing set consisting of 113 background-object pairs used in \cite{tarres2024thinking,song2024imprint}.

\noindent
\textbf{Evaluation Metrics} We assess alignment to input text prompt and object images via CLIP-Score \cite{hessel2021clipscore} (CLIP-T, CLIP-I) and DINO-Score \cite{oquab2023dinov2} (DINO). When bounding boxes are specified for each object, we compare each input object image to a corresponding cropped region within the generated image, as in \cite{chen2023anydoor,yang2023paintbyexample,tarres2024thinking}. In the customization setting, where models may not specify object locations, each input object is instead compared to the entire generated image, following \cite{li2024unimo,pan2023kosmos}. We refer to this metric variation as CLIP-I\textit{gl}, DINO\textit{gl}. For scenes with multiple objects, we average DINO(\textit{gl}) and CLIP-I(\textit{gl}) scores across all objects.
For compositing multiple objects with single-object compositing models \cite{yang2023paintbyexample,zhang2023controlcom,chen2023anydoor,song2024imprint, tarres2024thinking}, we add objects sequentially. Thus, we compute performance for all possible object-order sequences and average the results. To assess alignment between text prompts and composite results, we calculate CLIP-T by comparing the entire image to the text prompt (CLIP-T\textit{gl}) and by comparing the cropped compositing area to the text prompt (CLIP-T\textit{loc}), evaluating local and global alignment. Additionally, we conduct user studies to evaluate alignment to input modalities, quality of composited images, and realism of generated interactions.



\noindent
\textbf{Training Details} The U-Net and adaptor $\mathcal{A}$ are jointly trained on 8 A100, using Adam optimizer, learning rate of $4 \times 10^{-6}$ and effective batch size of 1024, via gradient accumulation. Losses are balanced using $\alpha = 10^{3}$, $\beta = 1$.

\subsection{Comparison to Existing Methods}

Our primary task is Object Compositing, but we also train for Subject-Driven Generation as an auxiliary task. Thus, we provide comprehensive evaluation across both tasks. 

\noindent
\textbf{Generative Object Compositing}
For this task, we compare to recent generative object compositing models: Paint by Example (PbE) \cite{yang2023paintbyexample}, ControlCom \cite{zhang2023controlcom}, AnyDoor \cite{chen2023anydoor}, IMPRINT \cite{song2024imprint}, Thinking Outside the BBox (TTOB) \cite{tarres2024thinking}. These models support compositing from a single object, background, and bounding box, requiring sequential runs to add multiple objects individually. For comparison, we evaluate our model in three additional inference modes: (i) without text guidance, (ii) sequential single-object compositing, (iii) sequential compositing without text guidance. 

\begin{figure*}[t]
    \centering
    \includegraphics[width=\linewidth]{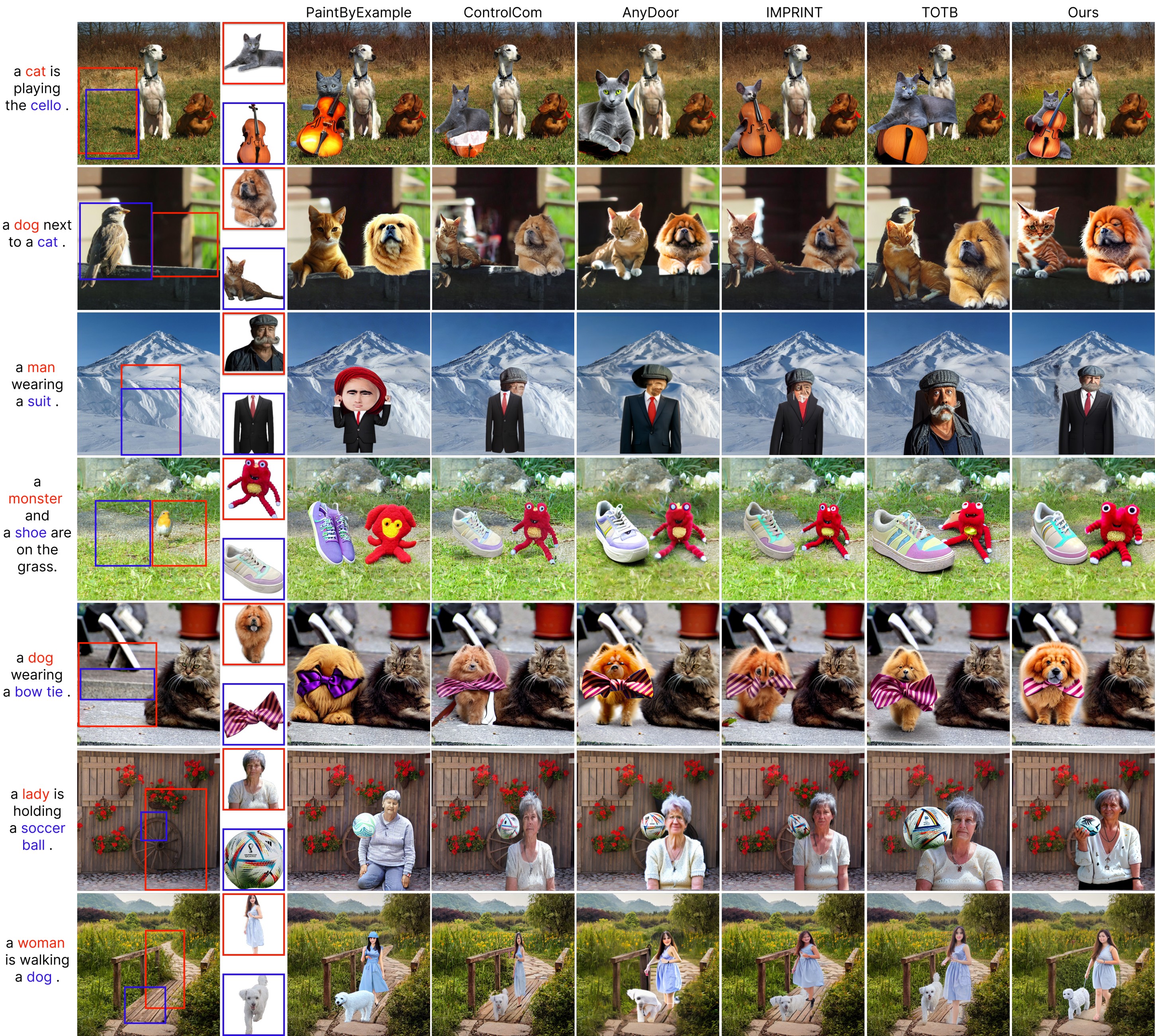}
    \caption{Visual comparison to Generative Object Compositing models \cite{yang2023paintbyexample,zhang2023controlcom,chen2023anydoor,song2024imprint,tarres2024thinking}. Our model provides more realistic, harmonious and natural-looking interaction between composited objects via simultaneous multi-object compositing. See SuppMat for more examples.}

    \label{fig:baselinescomp}
    \vspace{-6mm}
\end{figure*}

\begin{table}[t!]
\centering
\begin{adjustbox}{width=\linewidth}
\begin{tabular}{lcccccc}
\toprule
\multirow{2}[3]{*}{\textbf{Method}} & \multicolumn{2}{c}{\textbf{DreamBooth}} & \multicolumn{2}{c}{\textbf{MultiComp \textit{-overlap}}} & \multicolumn{2}{c}{\textbf{MultiComp \textit{-nonoverlap}}} \\
\cmidrule(lr){2-3} \cmidrule(lr){4-5} \cmidrule(lr){6-7}  & \textbf{CLIP-I$\uparrow$} & \textbf{DINO$\uparrow$} &  \textbf{CLIP-I$\uparrow$} & \textbf{DINO$\uparrow$} & \textbf{CLIP-I$\uparrow$} & \textbf{DINO$\uparrow$}\\ 
\cmidrule{1-7}
PbE \cite{yang2023paintbyexample}                                                                                                          &      0.778                   &  0.799             &        0.693             &   0.383  &     0.720 & 0.423                \\
\cmidrule{1-7}
ControlCom \cite{zhang2023controlcom}                                                                                          &      0.743          &  0.705 &          0.707                                                                 &                  0.478    &          0.740     &    0.543   \\ 
\cmidrule{1-7}
AnyDoor \cite{chen2023anydoor}                                                                                        &       0.806          & 0.836 &  0.727                                                               &       0.520          &   0.763    &               \textbf{0.593}         \\ 
\cmidrule{1-7}
IMPRINT \cite{song2024imprint}                                                                                                           &      \textbf{0.830}                   &  0.889   &           0.713                                                                                                         &        0.525             &  0.739   &      0.576                \\
\cmidrule{1-7}
TOTB \cite{tarres2024thinking}                                                                                            &    0.809      &  0.856   &      0.716    &                                                          0.485             &       0.740             &   0.531              \\ 
\cmidrule{1-7}
Ours                                                                                           &    0.803     &  0.892   &      \textbf{0.741}     &                                                              \textbf{0.532}             &   \textbf{0.768}            &   0.579              \\ 
\ \ w/o text                                                                                           &    0.816      &  \textbf{0.903}   &      0.729     &                                                          0.505             &       0.754             &    0.548             \\ 
\ \ sequential w/ text                                                                                          &    -      &  -   &      0.729     &   0.517                                                                  &       0.760             &    0.583              \\ 
\ \ sequential w/o text                                                                                           &    -      &  -   &      0.723     &    0.510                                                                &       0.756             &    0.578              \\

\bottomrule
\end{tabular}
\end{adjustbox}
\caption{Quantitative comparison of identity preservation against state-of-the-art generative object compositing methods. Single object compositing is evaluated on DreamBooth set. Two object compositing is on MultiComp set, comparing to both simultaneous and sequential uses of our model, by guiding inference with and without text. We distinguish between interacting (overlapping bboxes) and non-interacting cases for clarity. Details in SuppMat.} 
\label{tab:baselinescomp}
\vspace{-6mm}
\end{table}

As shown in Table \ref{tab:baselinescomp}, our model outperforms all compared models when composited objects interact, \ie when bounding boxes overlap. For non-overlapping or single-object compositions, our model maintains comparable identity and semantic preservation to state-of-the-art models. Furthermore, when objects overlap, compositing two objects simultaneously yields significantly better results than adding them sequentially, even with the same model, as in Fig \ref{fig:motivation}. Providing a text description of the interaction also boosts performance for scenes with multiple objects but is less critical when compositing a single object or non-interacting ones, where the expected composition is clearer.
Fig \ref{fig:baselinescomp} illustrates several advantages of compositing multiple objects simultaneously rather than sequentially: (i) it enables more cohesive harmonization and appearance consistency across objects and the scene (rows 1 - 5); (ii) it captures complex interactions involving object reposing with ease (rows 1, 6, 7); and (iii) with text guidance, the model can naturally complete scenes by adding any additional elements needed for realism (row 7).


\noindent
\textbf{Multi-Entity Subject-Driven Generation} 
We compare to customization methods accepting text and multiple object images as input (BLIP-Diffusion \cite{li2024blip}, KOSMOS-G \cite{pan2023kosmos}, UNIMO-G \cite{li2024unimo}). We also compare to Emu2Gen \cite{sun2024emugen}, additionally accepting layout guidance, therefore sharing input modalities with our model and allowing fair comparison.

\begin{table}[t!]

\centering
\begin{adjustbox}{width=\linewidth}

\begin{tabular}{lccccc}
\toprule
\textbf{Method}                            & \textbf{CLIP-I$\uparrow$} & \textbf{DINO$\uparrow$} & \textbf{CLIP-I\textit{gl}$\uparrow$} & \textbf{DINO\textit{gl}$\uparrow$} & \textbf{CLIP-T\textit{gl} $\uparrow$} \\ 
\midrule \multicolumn{6}{c}{Input: \textit{Text, Object Images}}
\\ \midrule
BLIP-Diffusion \cite{li2024blip}       &  - & -          &  0.675                                                                                                  &  0.455    &    0.249                        \\
\cmidrule{1-6}
KOSMOS-G \cite{pan2023kosmos}       &    - & -         & \textbf{0.704}                                                                                                  &  0.465      &   0.279       \\
\cmidrule{1-6}
UNIMO-G \cite{li2024unimo}  & - & -     &  0.699  &    \textbf{0.485}                                                                                                              &   0.293                     \\
\midrule \multicolumn{6}{c}{Input: \textit{Text, Object Images, Layout}} \\
\midrule
Emu2-Gen \cite{sun2024emugen}      &  0.595 & 0.414     &  0.616                                                                                                  &    0.434        &   0.287                                      \\
\cmidrule{1-6}
Ours    & \textbf{0.783} & \textbf{0.599}  &       0.688  &          0.454                                                                                                     &   \textbf{0.308}                             \\

\bottomrule
\end{tabular}
\end{adjustbox}

\caption{Quantitative comparison of identity preservation and text fidelity against multi-entity subject-driven generation methods. We compare to state-of-the-art methods on two-entity subset of MultiBench. For models with layout guidance, additional identity preservation metrics (CLIP-I, DINO) are provided by considering cropped regions around each object in the generated images.}
\label{tab:baselines2objcustomiz}
\vspace{-4mm}
\end{table}

\begin{figure}[t]
    \centering
    \includegraphics[width=\linewidth]{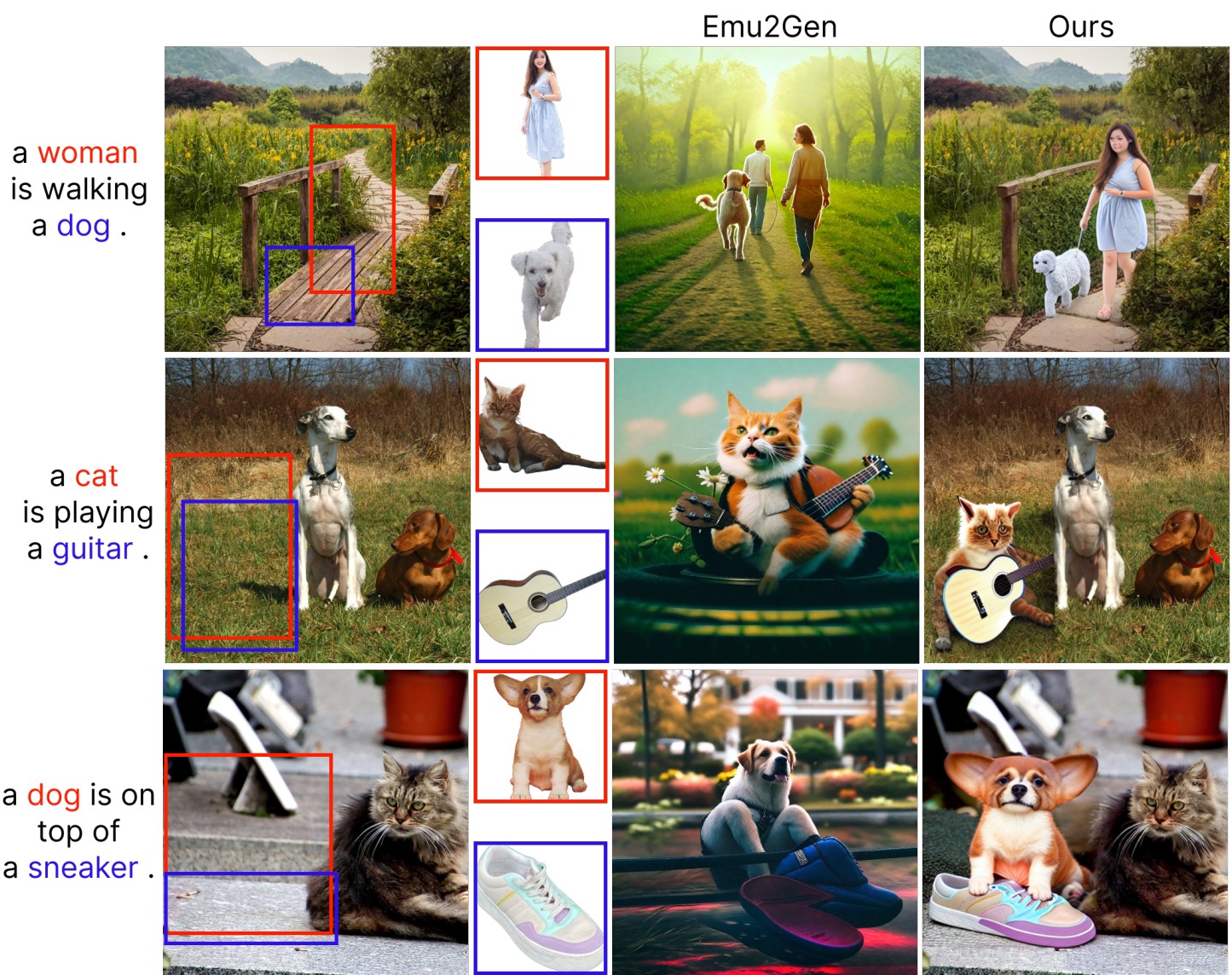}
    \caption{Visual comparison of our multi-object compositing model to Emu2Gen \cite{sun2024emugen}. For fair comparison, exact same set of inputs (including background image) is provided to each model. }

    \label{fig:baselineEmu}
    \vspace{-6mm}
\end{figure}

\noindent
Although customization is introduced as a proxy task to improve multi-object compositing, our model achieves comparable performance to state-of-the-art customization models, as shown in Table \ref{tab:baselines2objcustomiz}. Notably, our model obtains improved text alignment, benefiting from training data balancing local and global caption descriptions, as well as joint training with the complex task of object compositing. Additionally, our model demonstrates improved performance in patch-based image metrics (CLIP-I, DINO), indicating high layout alignment and identity preservation. In contrast, Emu2Gen obtains lower patch-based than global metrics. Refer to Fig 1 (bottom) and SuppMat for visual examples.



\begin{figure}[t]
    \centering
    \includegraphics[width=\linewidth]{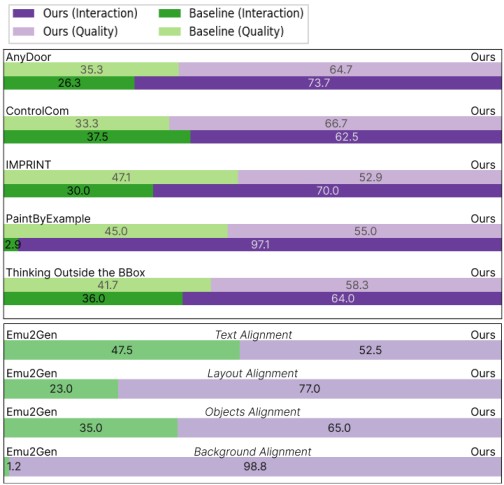}
    \caption{User Studies. \textit{Top:} Percentage of users prefering our method or each baseline \cite{chen2023anydoor,zhang2023controlcom,song2024imprint,yang2023paintbyexample,tarres2024thinking} on `image quality' and `realistic interaction'. \textit{Bottom}: Percentage of users prefering our method or Emu2Gen \cite{sun2024emugen} in terms of text, layout, objects and background alignment. All results are via majority consensus.} 

    \label{fig:userstudies}
    \vspace{-4mm}
\end{figure}

\noindent
\textbf{User Studies} We complement the comparison to State-of-Art models by completing various user studies (Fig \ref{fig:userstudies}). Non-expert users are presented with side-by-side generations from our model and each of the object compositing baselines, and are asked to choose their preferred image based on `most realistic interaction' (in cases of mask overlapping) and `best image quality'. Due to our model's ability to composite multiple interacting or non-interacting objects in a natural way, users prefer our model with up to 66.7\% preference in terms of image quality and up to 97.1\% preference for realistic interactions, via majority consensus. We also assess alignment to each individual input modality by comparing our compositing model to Emu2Gen \cite{sun2024emugen}. Emu2Gen is provided with the same inputs (Fig \ref{fig:baselineEmu}), presented as a multimodal input interlacing text, object images, object-specific bounding boxes and background. Background image is provided by adding \textit{`in $<\mathcal{I}_{BG}>$'} to the multimodal input, as showcased in \cite{sun2024emugen}. In this comparison, users prefer our alignment across all modalities. See SuppMat for details.



\subsection{Ablation Study}

\begin{figure}[t]
    \centering
    \includegraphics[width=\linewidth]{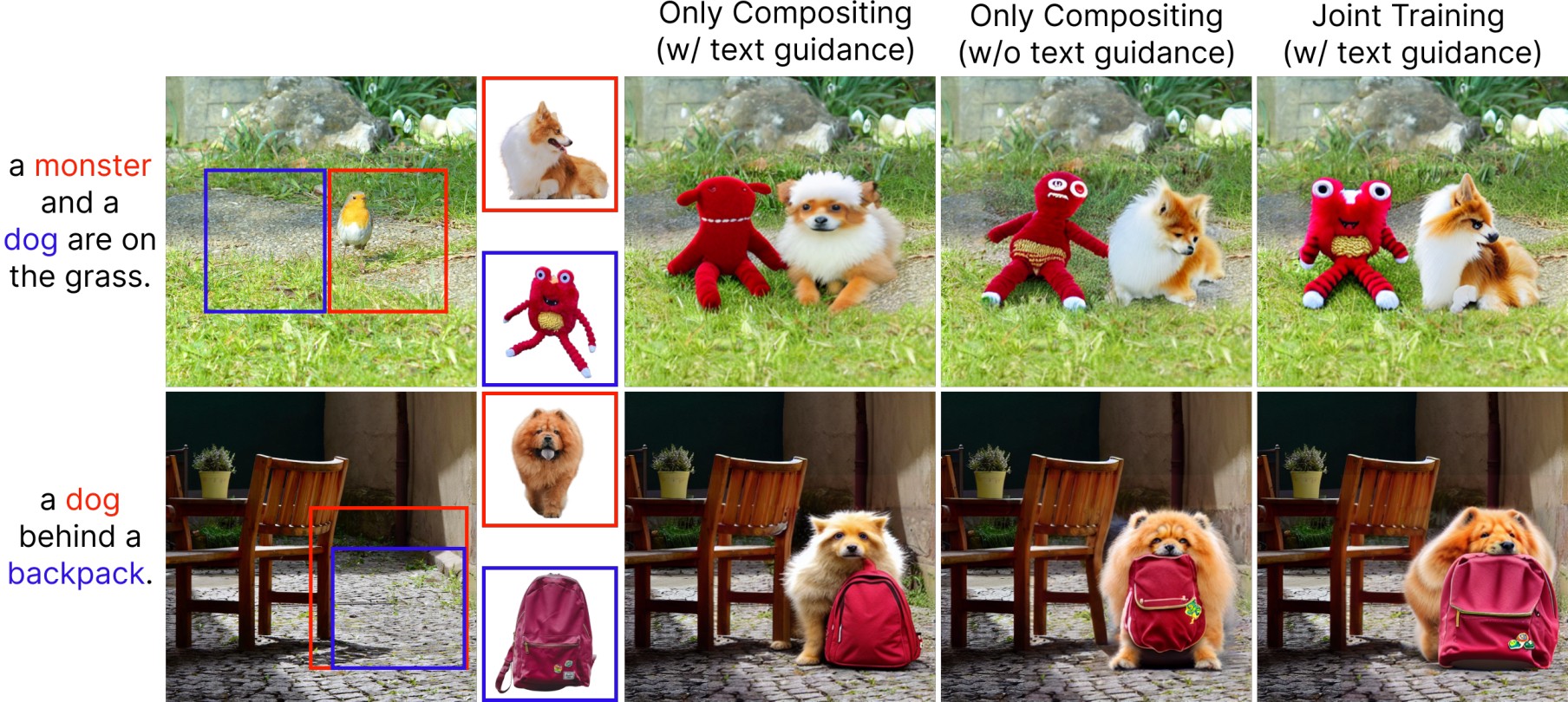}
    \caption{Visualization of the effect of Joint Training for Compositing and Customization tasks. When training solely on object compositing, balancing text and image alignment becomes significantly more complex. In this setup, the ablation model significantly drops identity preservation when inference is also guided with text. Our final model achieves better visual-language balance. }

    \label{fig:ablationcustom}
    \vspace{-6mm}
\end{figure}

\begin{figure}[t]
    \centering
    \includegraphics[width=\linewidth]{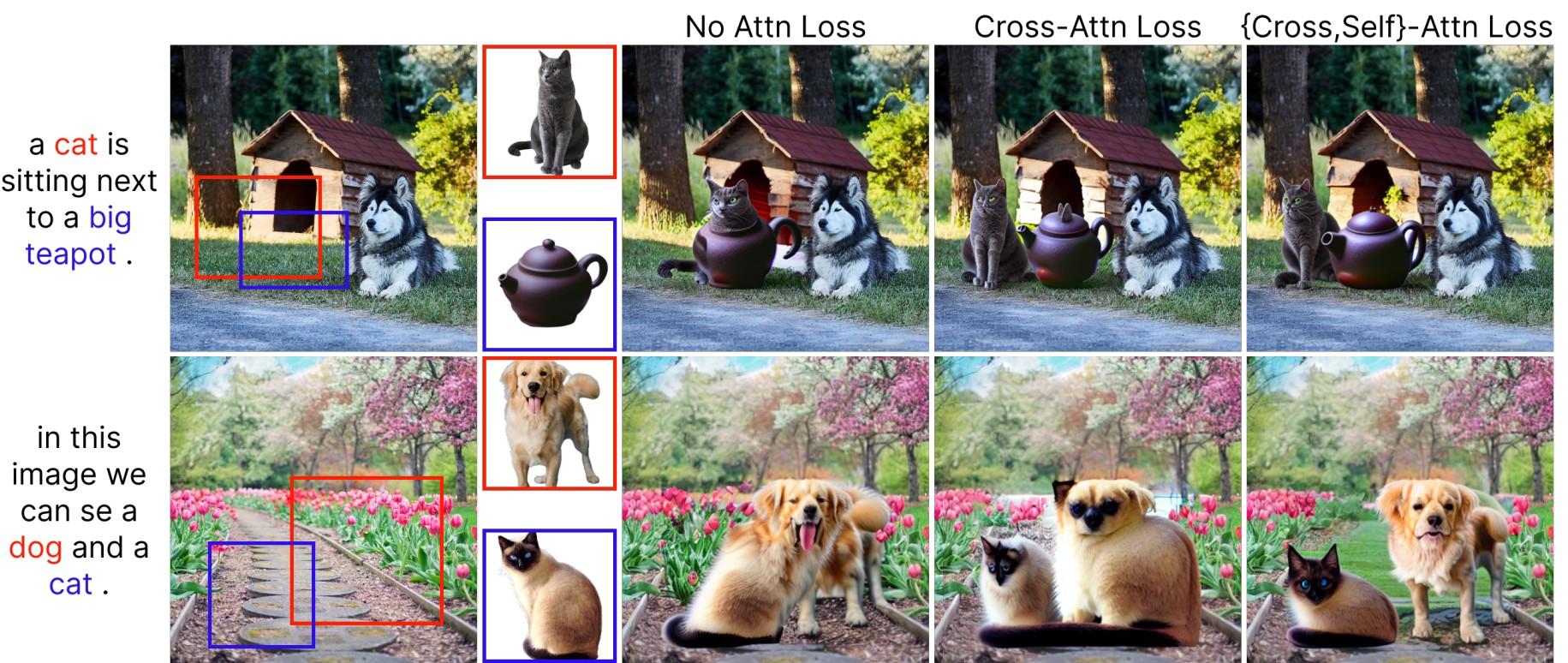}
    \caption{Visualization of the effect of $\mathcal{L}_c$, $\mathcal{L}_s$. Without attention-based losses, our model merges objects with similar semantics or visual traits. Cross-attention loss $\mathcal{L}_c$ improves identity separation, but some leakage remains (\eg, cat ears on dog and teapot). Adding self-attention loss $\mathcal{L}_s$ further reduces this leakage.} 

    \label{fig:ablationloss}
    \vspace{-2mm}
\end{figure}

\begin{table}[t!]

\centering
\begin{adjustbox}{width=0.47\textwidth}

\begin{tabular}{lllll} 
\toprule
\textbf{Method}                            & \textbf{DINO$\uparrow$} & \textbf{CLIP-I$\uparrow$} & \textbf{CLIP-T\textit{loc}$\uparrow$} & \textbf{CLIP-T\textit{gl}$\uparrow$} \\ 
\midrule \multicolumn{5}{c}{MultiComp-\textit{action}}
\\ \midrule
Ours       &    \textbf{0.540}                                                                                                  &  0.745              &   0.286   &                    0.285                      \\
\ \ w/o self loss       &    0.538                                                                                                  &  0.744              &   0.283   &                    0.284                    \\
\ \ w/o attn loss       &    0.534                                                                                                  &  0.739             &   0.270   &                    0.274                     \\
\ \ w/o customiz.       &    0.449                                                                                                 &  0.705              &   \textbf{0.295}   &                    \textbf{0.298}                     \\
\ \ w/o multi-view       &    0.535                                                                                                  &  \textbf{0.751}              &   0.268   &                    0.273                     \\
\midrule \multicolumn{5}{c}{MultiComp-\textit{positional}}
\\ \midrule
Ours       &    0.585                                                                                                  &  0.773              &   0.292   &                    0.294                      \\
\ \ w/o self loss       &    0.584                                                                                                  &  0.770              &   0.289   &                    0.291                     \\
\ \ w/o attn loss       &    0.581                                                                                                  &  0.770             &   0.281   &                    0.285                     \\
\ \ w/o customiz.       &    0.474                                                                                                 &  0.734              &   \textbf{0.298}   &                    \textbf{0.298}                     \\
\ \ w/o multi-view       &    \textbf{0.591}                                                                                                  &  \textbf{0.780}              &   0.280   &                    0.283                     \\

\bottomrule
\end{tabular}
\end{adjustbox}

\caption{Quantitative comparison of text and image alignment to different ablations of our compositing model. For better analysis we distinguish between \textit{action} and \textit{positional} subsets of MultiComp set, based on objects relationship in text description. We compare our full model to different versions obtained without (i) self-attention loss $\mathcal{L}_s$; (ii) self and cross-attention losses ($\mathcal{L}_c$,$\mathcal{L}_s$); (iii) joint training for compositing and customization; and (iv) multi-view data (\ie video, manually collected set).}
\label{tab:ablations}
\vspace{-6mm}
\end{table}

Table \ref{tab:ablations} displays identity and text fidelity metrics for various ablations of our model. (i) As seen in Fig \ref{fig:ablationloss}, adding attention-based losses helps prevent semantic and visual information leakage between objects, improving both identity preservation and text alignment, as well as the overall image quality. (ii) When training solely for object compositing, the model must learn to add new objects to a scene, repose them according to actions described in text input, complete the scene by adhering to the same text, and harmonize and relight the image to make everything look seamless while preserving object identity. By introducing customization as an auxiliary task and alternating between both tasks during training, the model can sometimes relieve the inpainting aspect of compositing, focusing only on balancing text alignment and identity preservation. As in Fig \ref{fig:ablationcustom}, in the absence of the customization task, the model struggles to balance text and image preservation, and the objects become irrecognizable. (iii) Without multi-view data, even if visual and geometrical transformations are applied to the object images, the model lacks the ability to repose them to match the actions described in the text. This results in a `copy-paste' behavior where objects are minimally transformed (high CLIP-I, DINO), preserving identity but leading to highly unnatural results and low text alignment (CLIP-T\textit{gl},\textit{loc}).


\vspace{-2mm}
\subsection{Applications}

Although not explicitly trained for them, our model exhibits some emerging capabilities. See SuppMat for more.

\begin{figure}[t]
    \centering
    \includegraphics[width=\linewidth]{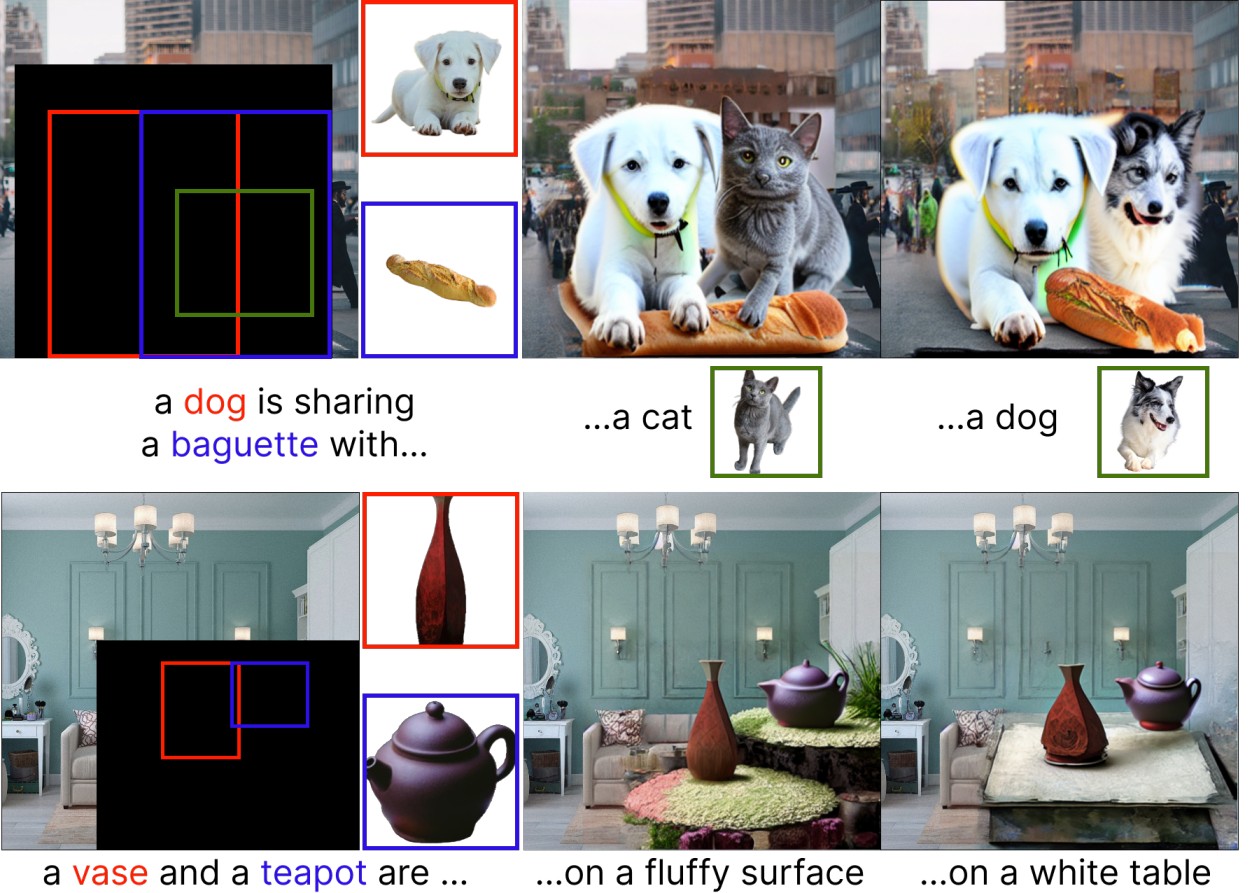}
    \caption{Applications of our model. \textit{Top:} Simultaneous Multi-Object Compositing of Three Objects. \textit{Bottom:} Subject-Driven Inpainting. Our model is able of simultaneously compositing objects while completing the scene with text and image alignment.} 

    \label{fig:applications}
    \vspace{-6mm}
\end{figure}

\textbf{Multi-Object Generation} As shown in Fig \ref{fig:applications} (top), our model is able to perform multi-object compositing with more than two objects. By learning one-to-one object interactions, it develops a strong prior that allows it to generalize to compositing multiple objects simultaneously. 

\textbf{Subject-Driven Inpainting} Our joint training enables the model to learn key subtasks, such as background synthesis, blending, harmonization, and reposing. As shown in Fig \ref{fig:applications} (bottom), these skills can be applied to subject-driven inpainting. In this task, the model uses text and layout guidance to seamlessly complete a scene, generating and integrating additional objects while maintaining a natural and coherent composition with the given visuals. 


\vspace{-2mm}
\subsection{Limitations}

Although our model can handle two or more objects, the pipeline is not designed for an unlimited number. As more objects are added, the multimodal embedding grows, which could lead to scalability issues. A potential future improvement could involve feeding each object embedding separately, impacting only corresponding attention maps. Additionally, we base our model on SD1.5 due to compute limitation. Overall harmonization and composition quality could be further improved with stronger diffusion base model such as SDXL \cite{podell2023sdxl} or SD3 \cite{esser2024scaling}.



\vspace{-2mm}
\section{Conclusion}
\label{sec:conclusion}
\vspace{-1mm}
In this paper, we introduce the first generative model capable of simultaneously compositing multiple objects in a natural and realistic manner, guided by text and layout. We demonstrate how joint training on object compositing and subject-driven generation improves model performance, achieving state-of-the-art results in both tasks. This multi-object, dual training approach enhances flexibility, enabling applications like subject-driven inpainting and more realistic object interactions. We hope this work inspires further research into more controllable and effective object compositing techniques.

{
    \small
    \bibliographystyle{ieeenat_fullname}
    \bibliography{main}
}

\clearpage
\maketitlesupplementary
\setcounter{page}{1}
\setcounter{figure}{0}
\setcounter{section}{0}

\section{Training and Testing Data}

In this section, we provide additional information about the training data generation pipeline proposed in Main Paper Section \ref{sec:datageneration} and MultiComp, the multi-object compositing test set introduced in Main Paper Section \ref{sec:experiments}.

\subsection{Training Data}

As detailed in Main Paper Section \ref{sec:datageneration}, our paired training data --- comprising ground truth images with multiple objects, descriptive captions with grounding information, segmented images of two objects, and corresponding object-specific bounding boxes --- is collected from complementary sources: Video Data, In-the-Wild Images, and Manually Collected Data. Below, we expand on how paired data is extracted from each source.

\begin{figure}[t]
    \centering
    \includegraphics[width=\linewidth]{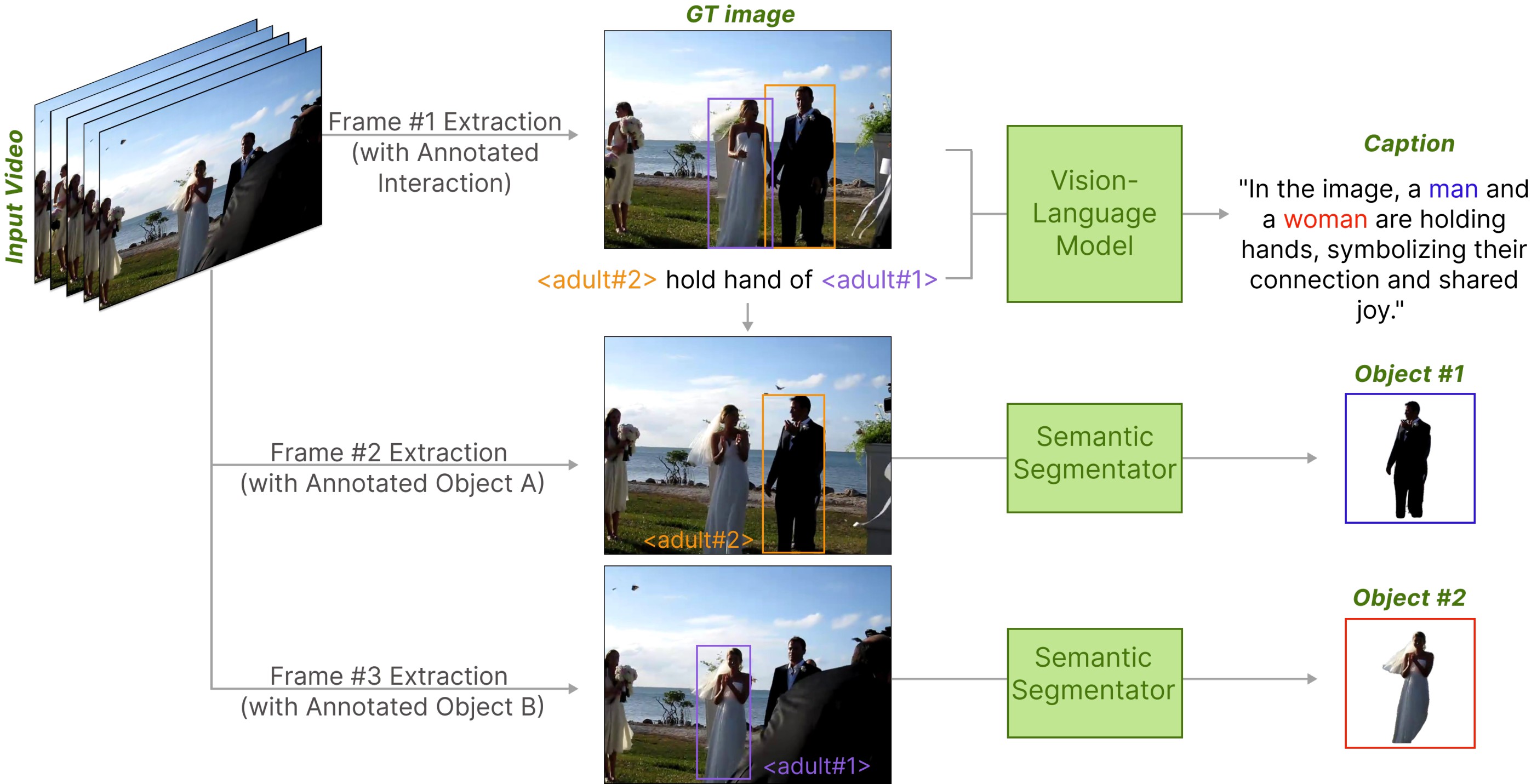}
    \caption{Training Data Generation from Video Data. Paired training data is obtained from video object relation datasets \cite{shang2017video,shang2019annotating} by extracting three frames with corresponding annotations and leveraging Vision-Language Models \cite{liu2023improvedllava}.}
    \label{fig:suppvideo}
\end{figure}

\noindent\textbf{Video Data} Fig \ref{fig:suppvideo} illustrates the process for extracting paired training data from videos in \cite{shang2017video, shang2019annotating}. A ground-truth frame containing an annotated relationship between two objects is randomly selected. Two additional frames, each showing one of the interacting objects, are extracted from the same video, ensuring a similarity between the views of each object (DINO score MSE $\geq 0.8$) \cite{oquab2023dinov2}. A caption describing the relationship is automatically generated by feeding the ground-truth image and annotated relation into LLaVA v1.6 (34B) \cite{liu2023improvedllava} with the prompt:
\textit{``Can you provide a grammatically correct one-line caption for the relation $<$object A$>$ $<$relation$>$ $<$object B$>$ in the image?"}.
The segmented objects are then extracted using an off-the-shelf semantic segmentation model \cite{qi2022entityseg}.


\noindent\textbf{Image Data} We propose two automated approaches for obtaining paired data from in-the-wild images. 

\begin{figure}[t]
    \centering
    \includegraphics[width=\linewidth]{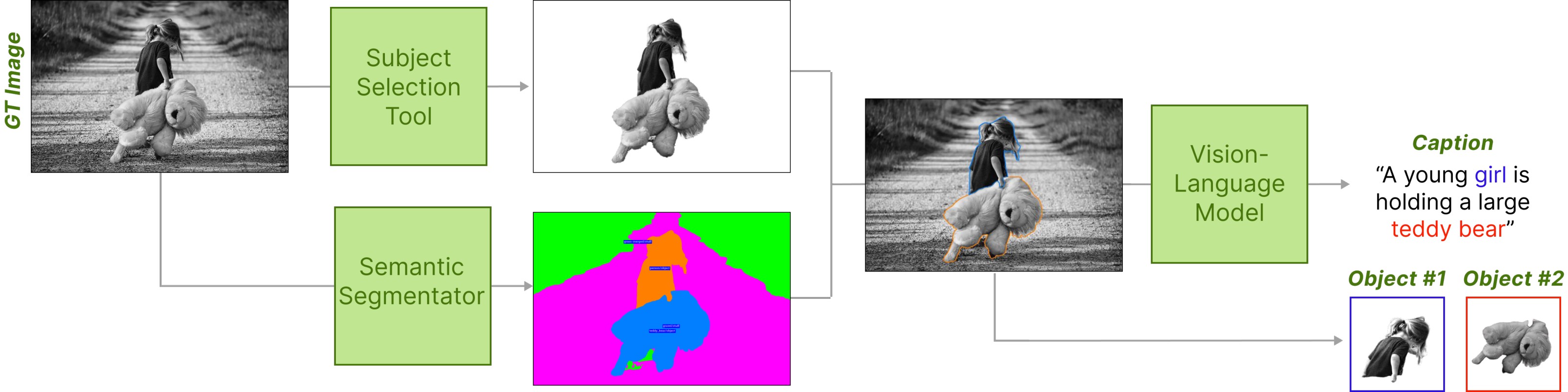}
    \caption{Training Data Generation from Image Data via Top-Down Approach. Paired training data is derived from in-the-wild images by leveraging a Vision-Language Model \cite{cai2024vip} and a Semantic Segmentator \cite{qi2022entityseg}.}
    \label{fig:suppimagevip}
\end{figure}

\textbf{Top-down approach} As illustrated in Fig \ref{fig:suppimagevip}, this approach generates paired training data from a single image using a systematic process. First, a commercial subject selection tool identifies the main objects in the scene. A semantic segmentation model \cite{qi2022entityseg} then segments the selected region to determine the number of objects present. If multiple objects are detected, two are randomly selected as composited objects. Their outlines are highlighted on the image using distinct colors (e.g., orange and blue) and passed to ViP-LLaVA (13B) \cite{cai2024vip} for caption generation through two sequential prompts. In the first step, entities within each highlighted outline are identified with a question like: \textit{``Please follow the sentence pattern of the example to list the entities within each rectangle. Example: `orange: banana; blue: apple'"}. This produces responses such as \textit{``orange: teddy bear; blue: girl"}. Using these entity labels, the second step generates a descriptive caption with a prompt like: \textit{``Can you provide a one-line caption including the interaction between `teddy bear within the orange rectangle' and `girl within the blue rectangle' in the image by using these exact entity names?"}. The resulting caption, \eg, \textit{``A young girl within the blue rectangle is holding a large teddy bear within the orange rectangle"}, is refined by removing the grounding phrases \textit{``within the orange/blue rectangle"} after using them for correlating object images with text tokens. This method ensures accurate grounding information, even when both entities are labeled the same, resulting in the final caption: \textit{``A young girl is holding a large teddy bear"}.

\begin{figure}[t]
    \centering
    \includegraphics[width=\linewidth]{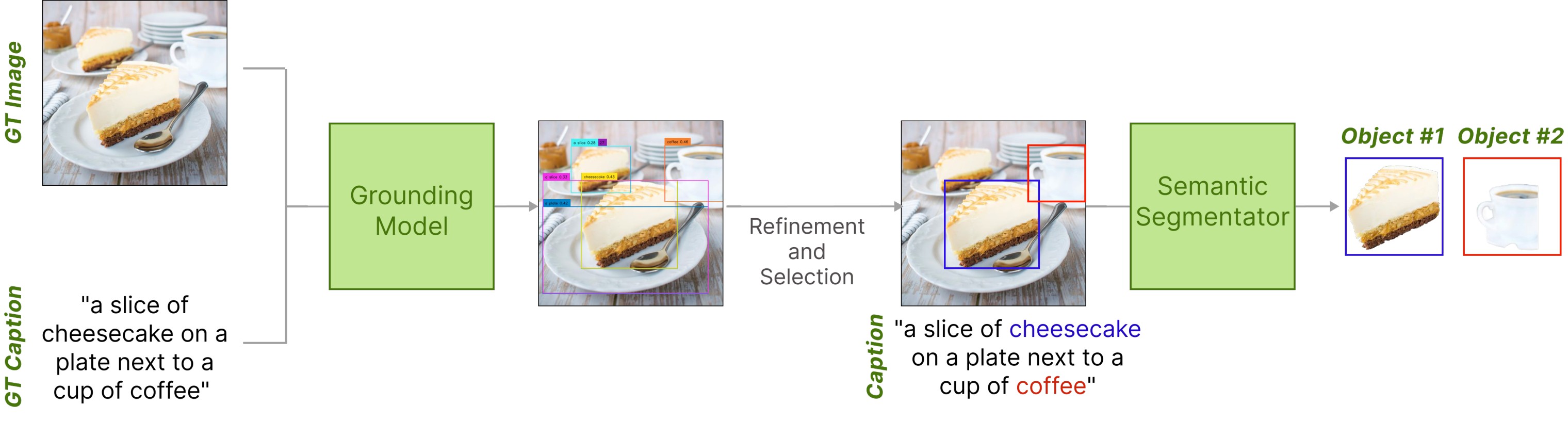}
    \caption{Training Data Generation from Image Data via Bottom-Up Approach. Paired training data is extracted from in-the-wild images with a paired caption by leveraging a Grounding Model \cite{liu2023groundingdino} and a Semantic Segmentator \cite{qi2022entityseg}.}
    \label{fig:suppimagedino}
\end{figure}


\textbf{Bottom-up approach} This approach (Fig \ref{fig:suppimagedino}) leverages a grounding model like GroundingDINO \cite{liu2023groundingdino} to process paired ground truth images and captions. The model extracts bounding boxes that link specific words in the caption to objects in the image. Duplicates, background elements, and undesired objects (\eg, overly large or small objects, or those with low confidence scores) are removed, leaving a set of object candidates with corresponding grounding information. Two of those objects are then randomly selected, and an off-the-shelf semantic segmentation model \cite{qi2022entityseg} is used to extract them. This results in two segmented objects along with their associated grounding details from the original caption.

\begin{figure}[t]
    \centering
    \includegraphics[width=\linewidth]{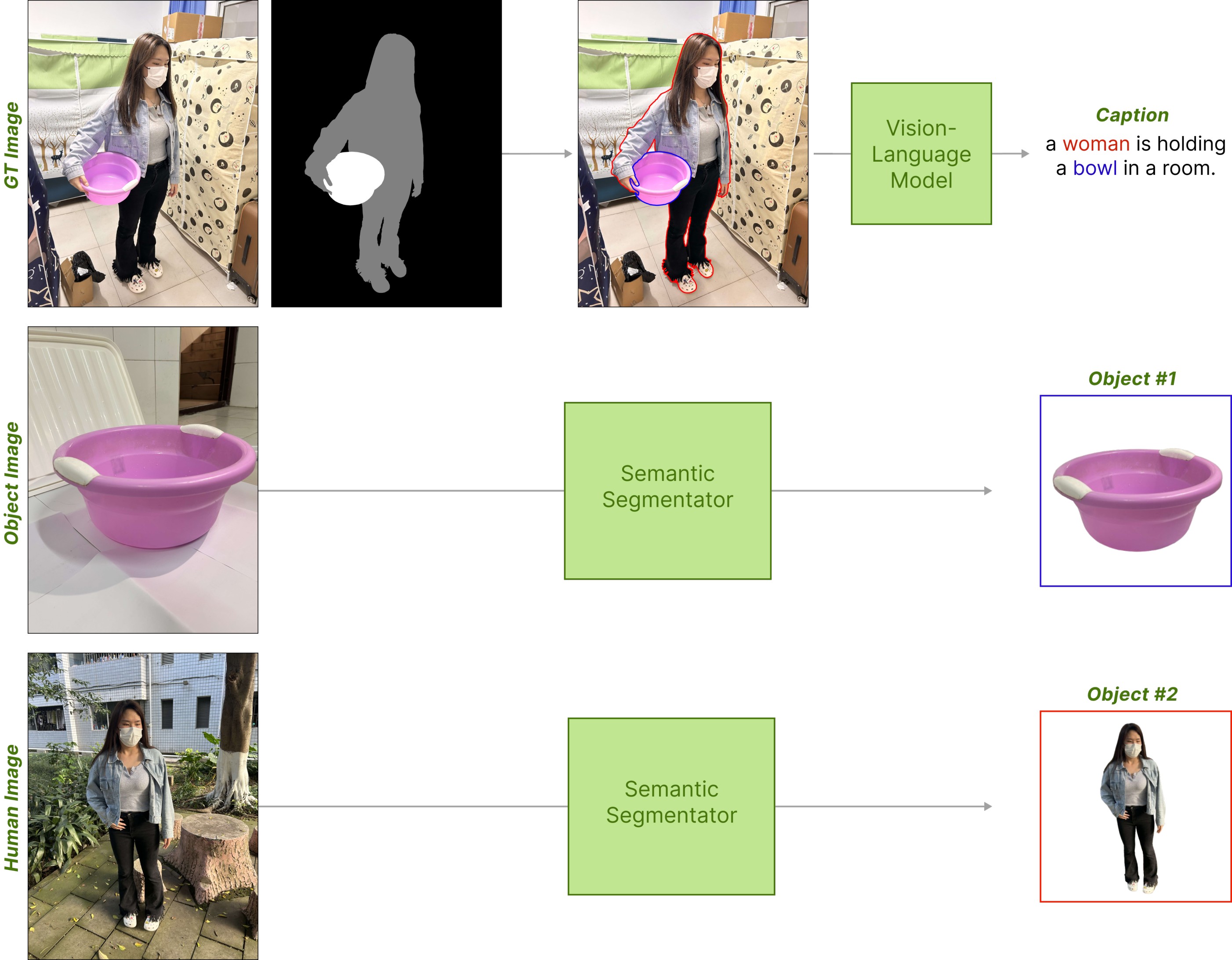}
    \caption{Training Data Generation from Manually Collected Data. Paired training data is obtained from a collected dataset containing object images, human images, and images of humans interacting with objects. We leverage a Vision-Language Model \cite{cai2024vip} and a Semantic Segmentator \cite{qi2022entityseg} to extract segmented objects and corresponding caption with grounding information.}
    \label{fig:suppcollected}
\end{figure}

\noindent\textbf{Manually Collected Data} We manually collect and annotate images featuring objects, humans, and human-object interactions. For captioning with grounding information, we use ViP-LLaVA (13B) \cite{cai2024vip}, following the exact same procedure as the top-down approach described above. Additionally, a semantic segmentation model \cite{qi2022entityseg} is employed to segment entities in images containing either a single object or a human. Fig \ref{fig:suppcollected} illustrates this approach.


\section{Inference Data}


Our collected MultiComp set consists of 119 paired data entries, each containing: (i) a background image, (ii) two object images, (iii) object-specific bounding boxes, (iv) an inpainting bounding box encompassing the previous ones, and (v) a descriptive caption with grounding information. Background images are sourced from Pixabay \cite{pixabay}, while objects are from Pixabay \cite{pixabay}, MultiBench \cite{li2024unimo}, and DreamBooth \cite{ruiz2023dreambooth}. Bounding boxes and captions are manually crafted. For evaluation, we perform 5 iterations of each model on the entire set, resulting in 595 generated images.

Out of these, 395 images contain overlapping bounding boxes for the two objects (MultiComp-\textit{overlap}), while 200 show non-overlapping bounding boxes (MultiComp-\textit{nonoverlap}). We evaluate these subgroups separately due to their differing levels of difficulty. While simultaneous compositing offers benefits like cohesive harmonization in both cases, it is especially effective in the overlapping cases, where it allows for simultaneous object reposing and the generation of additional elements needed for the scene. 


When textual input is provided, we further categorize the set into two subgroups based on caption types: MultiComp-\textit{action} and MultiComp-\textit{positional}. The MultiComp-\textit{action} subset contains 375 images generated from action-based captions (\eg, \textit{running after, playing with, holding}), while the MultiComp-\textit{positional} subset includes 220 images generated from captions describing positional relations (\eg, \textit{next to, behind, in front}). This distinction allows us to separately evaluate cases where reposing objects is often necessary (MultiComp-\textit{action}) versus those primarily describing object layout (MultiComp-\textit{positional}).


\section{Comparison to Existing Methods}

We provide additional visualizations comparing our model to existing generative object compositing and multi-entity subject-driven generation models in Sections \ref{sec:supp_compbaselines} and \ref{sec:supp_custombaselines}. Further details on user studies can be found in Section \ref{sec:supp_userstudies}.

\subsection{Comparison to Generative Object Compositing Methods}
\label{sec:supp_compbaselines}

\begin{figure*}[t]
    \centering
    \includegraphics[width=0.98\linewidth]{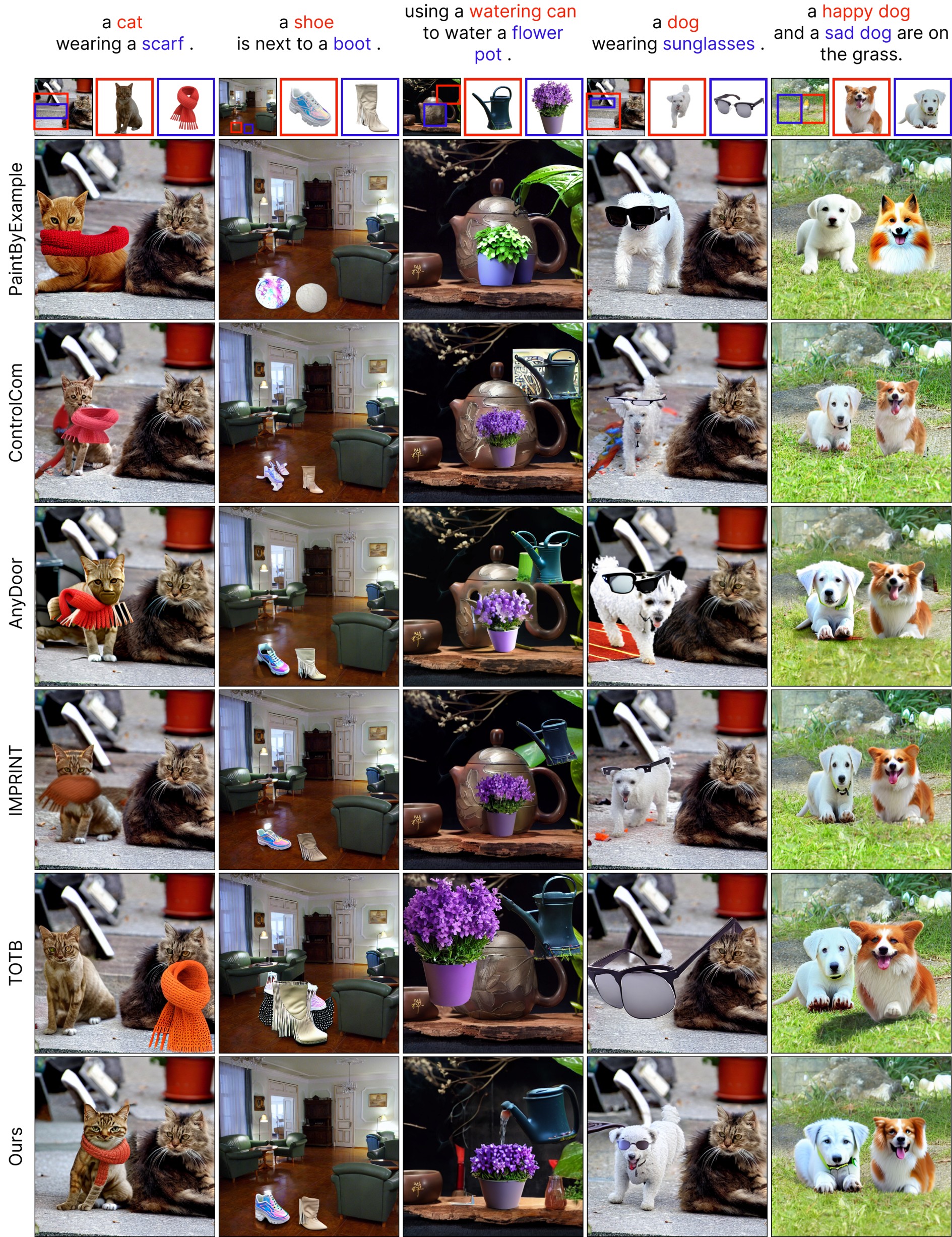}
    \caption{Visual comparison of our Multi-Object Compositing Method and State-of-the-Art Generative Object Compositing Methods \cite{yang2023paintbyexample,zhang2023controlcom,chen2023anydoor,song2024imprint,tarres2024thinking}.}

    \label{fig:supp_baselinescomp}
\end{figure*}

We visually compare our simultaneous multi-object compositing method to sequentially adding two objects using State-of-the-Art Generative Compositing Methods \cite{yang2023paintbyexample,zhang2023controlcom,chen2023anydoor,song2024imprint,tarres2024thinking} in Fig \ref{fig:supp_baselinescomp}.

\subsection{Comparison to Subject-Driven Generation Methods}
\label{sec:supp_custombaselines}

\begin{figure*}[t]
    \centering
    \includegraphics[width=\linewidth]{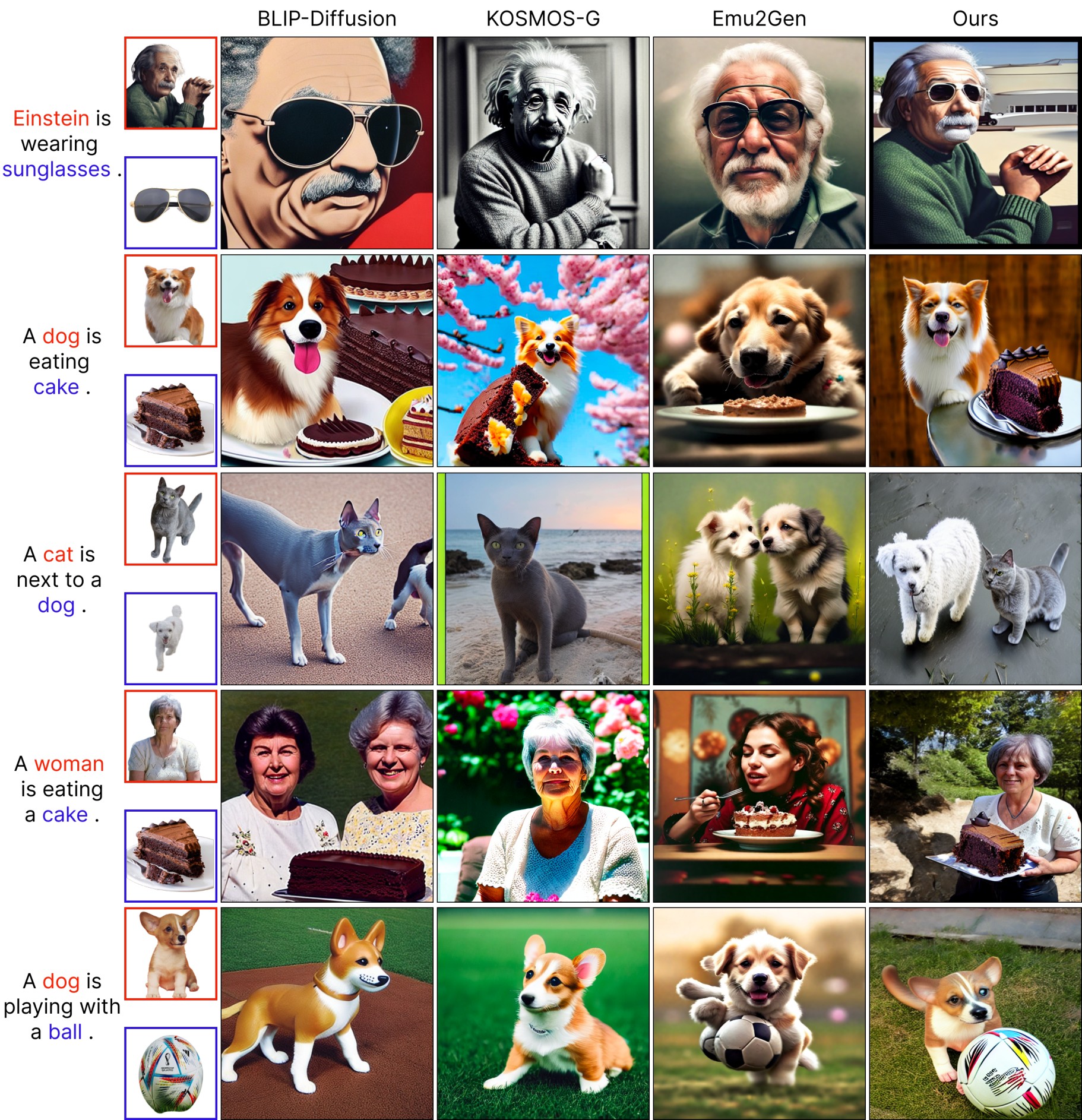}
    \caption{Visual comparison of our Customization Method and State-of-the-Art Subject-Driven Generation Methods \cite{li2024blip,pan2023kosmos,sun2024emugen}.}

    \label{fig:supp_baselinescustom}
\end{figure*}

We visually compare two-entity subject-driven generation using our method and existing methods with available code (BLIP-Diffusion \cite{li2024blip}, KOSMOS-G \cite{pan2023kosmos} and Emu2Gen \cite{sun2024emugen}) in Fig \ref{fig:supp_baselinescustom}.

\subsection{User Studies}
\label{sec:supp_userstudies}

\begin{figure}[t]
    \centering
    \includegraphics[width=\linewidth]{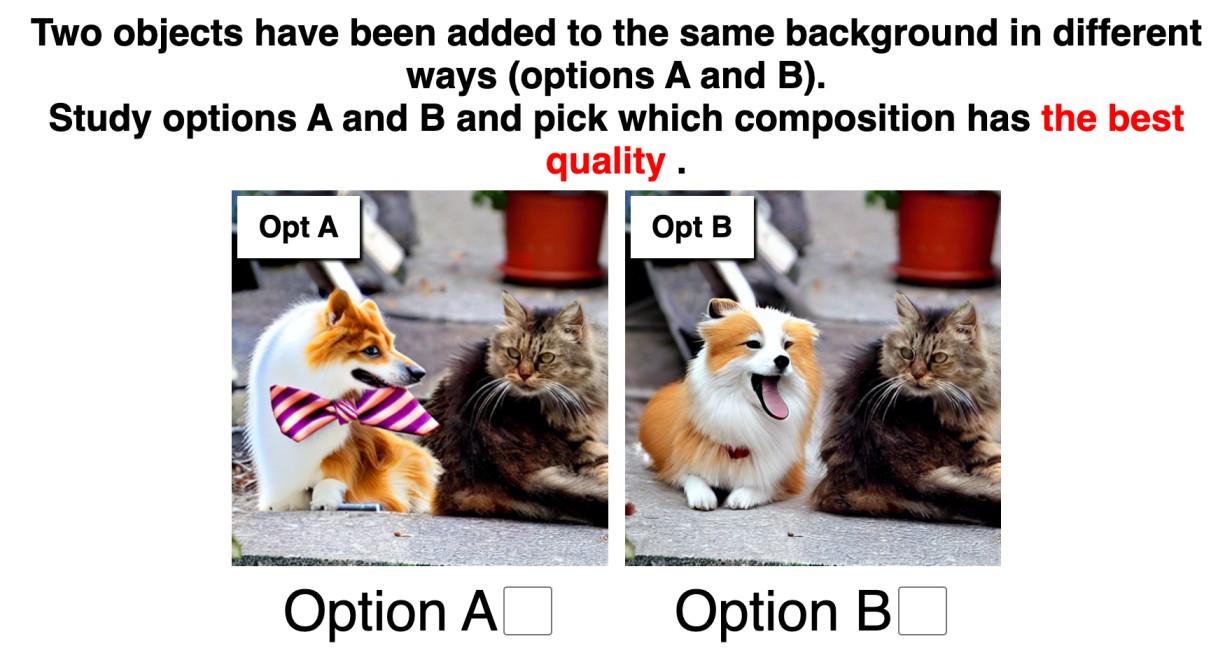}
    \caption{User Study on `Compositing Quality'. Screenshot of user study presented to participants for evaluating the image quality of our multi-object compositing method against generative object compositing baselines \cite{yang2023paintbyexample,zhang2023controlcom,chen2023anydoor,song2024imprint,tarres2024thinking}.}

    \label{fig:supp_userquality}
\end{figure}

\begin{figure}[t]
    \centering
    \includegraphics[width=\linewidth]{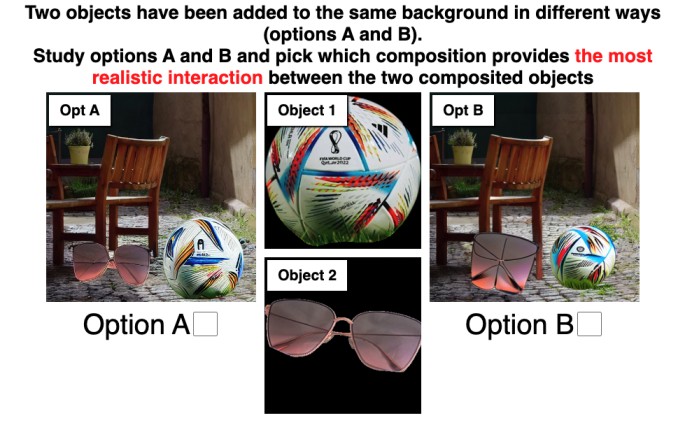}
    \caption{User Study on `Realistic Interaction'. Screenshot of user study presented to participants for evaluating the realism of interactions generated by our multi-object compositing method against generative object compositing baselines \cite{yang2023paintbyexample,zhang2023controlcom,chen2023anydoor,song2024imprint,tarres2024thinking}.}

    \label{fig:supp_userinteraction}
\end{figure}

\begin{figure}[t]
    \centering
    \includegraphics[width=\linewidth]{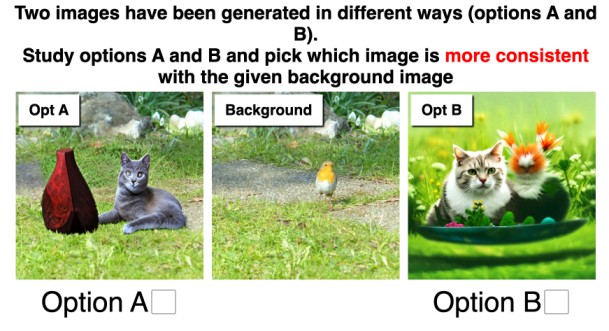}
    \caption{User Study on `Background Alignment'. Screenshot of user study presented to participants for evaluating the alignment with background image of our multi-object compositing method against Emu2Gen \cite{sun2024emugen}.}

    \label{fig:supp_userbg}
\end{figure}

\begin{figure}[t]
    \centering
    \includegraphics[width=\linewidth]{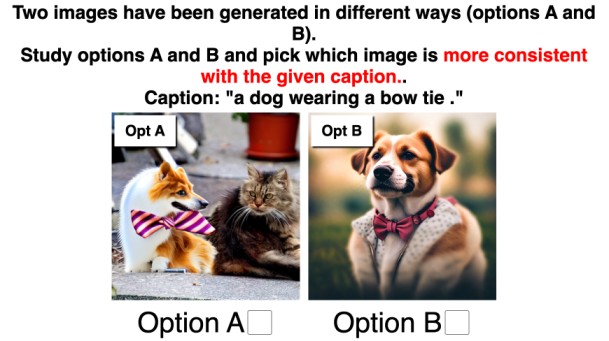}
    \caption{User Study on `Text Alignment'. Screenshot of user study presented to participants for evaluating the text alignment of our multi-object compositing method against Emu2Gen \cite{sun2024emugen}.}

    \label{fig:supp_usercaption}
\end{figure}

\begin{figure}[t]
    \centering
    \includegraphics[width=\linewidth]{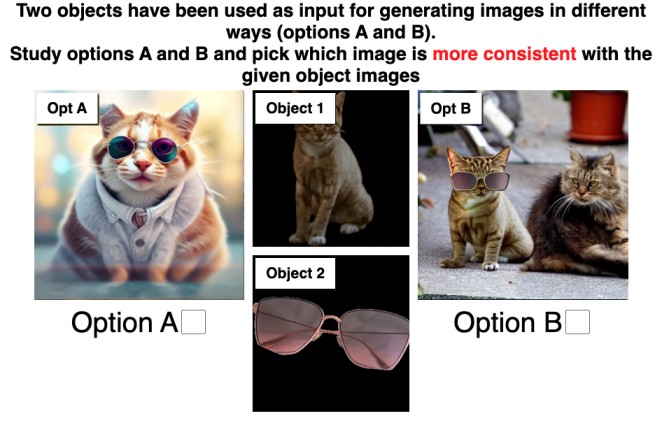}
    \caption{User Study on `Objects Alignment'. Screenshot of user study presented to participants for evaluating the alignment with input object images of our multi-object compositing method against Emu2Gen \cite{sun2024emugen}.}

    \label{fig:supp_userobjs}
\end{figure}

\begin{figure}[t]
    \centering
    \includegraphics[width=\linewidth]{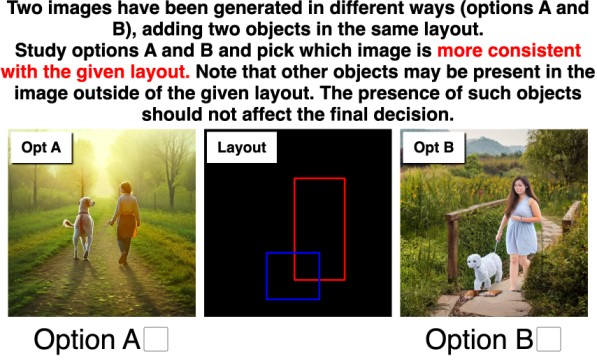}
    \caption{User Study on `Layout Alignment'. Screenshot of user study presented to participants for evaluating the layout alignment of our multi-object compositing method against Emu2Gen \cite{sun2024emugen}.}

    \label{fig:supp_userlayout}
\end{figure}

We conduct six user studies to evaluate our multi-object compositing model against other generative object compositing models \cite{yang2023paintbyexample,zhang2023controlcom,chen2023anydoor,song2024imprint,tarres2024thinking} and Emu2Gen \cite{sun2024emugen}. In each study, non-expert users are shown two side-by-side images, one generated by our model and the other by a baseline, presented in random order. Users are asked to choose the preferred image based on a specific criterion. The entire MultiComp set is used for each experiment, except for the `most realistic interaction' evaluation, where only images from MultiComp-\textit{overlap} are considered, as this subset best evaluates the task. At least five users rate each image pair, and the results are aggregated via majority consensus. Visual examples of each experiment and the specific questions posed to the users can be found in Figs \ref{fig:supp_userquality}, \ref{fig:supp_userinteraction}, \ref{fig:supp_userbg}, \ref{fig:supp_usercaption}, \ref{fig:supp_userobjs}, and \ref{fig:supp_userlayout}.



\section{Ablation Study}

\begin{figure*}[t]
    \centering
    \includegraphics[width=\linewidth]{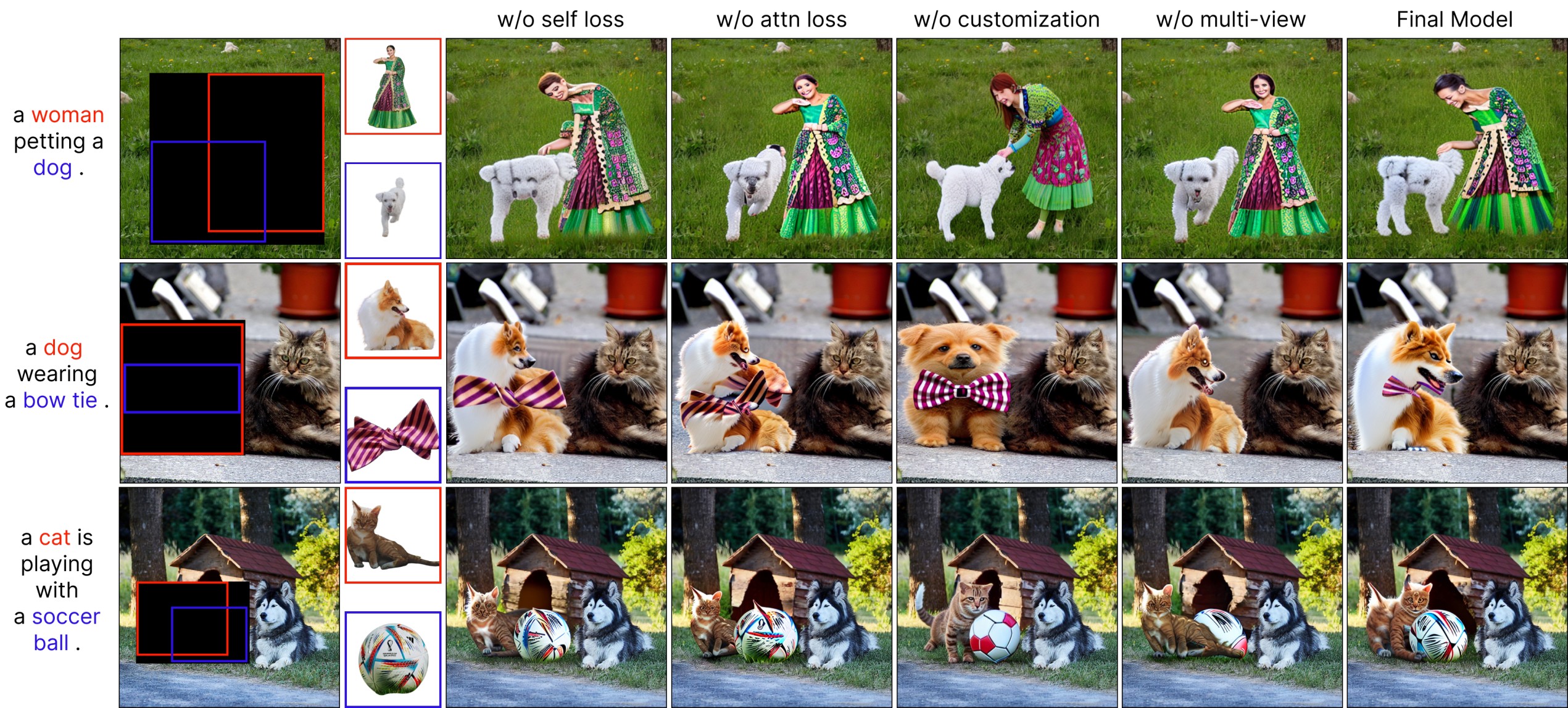}
    \caption{Visual examples for each ablation of the model. From left to right: (i) inputs (background, layout, objects and text), (ii) no self-attention loss, (iii) no self-attention or cross-attention loss, (iv) no joint training for compositing and customization, (v) no multi-view data (\ie video data, manually collected data), (vi) final model.}

    \label{fig:supp_ablation}
\end{figure*}

Fig \ref{fig:supp_ablation} shows visual examples of images generated by each ablation of our model (as detailed in Main Paper Table \ref{tab:ablations}) for the same set of inputs. Without multi-view data (\ie, video data, manually collected data), the model struggles to properly repose and combine objects to align with the textual description. In the absence of joint training for compositing and customization, the model fails to balance textual and visual inputs, resulting in object identity loss when reposing. Lastly, without cross-attention and/or self-attention losses, disentangling object identities becomes difficult, leading to texture and color leakage between objects (\eg bow color, cat ear on ball).


\section{Applications}
\subsection{Model versatility}

\begin{figure*}[t]
    \centering
    \includegraphics[width=\linewidth]{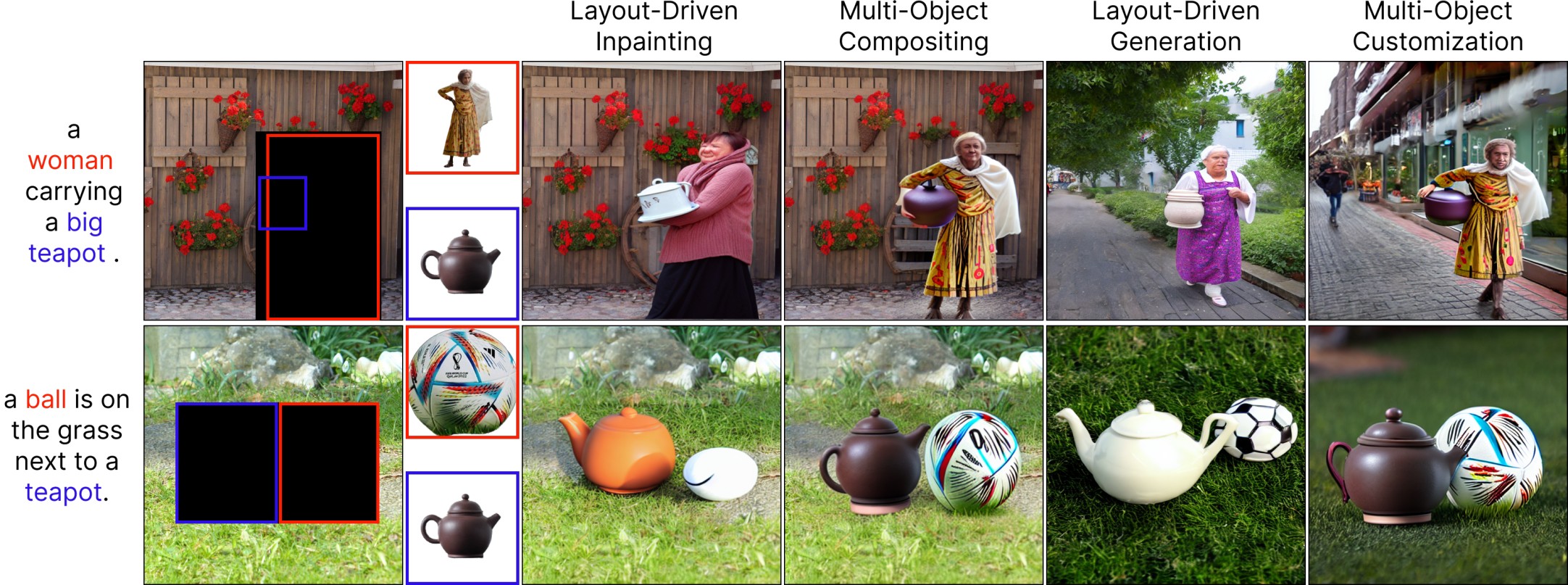}
    \caption{Visual examples for different applications of our model. Our model can operate on different modes such as: (i) layout-driven inpainting, (ii) multi-object compositing, (iii) layout-driven generation, (iv) multi-entity subject-driven generation.}
    \label{fig:supp_versatility}
\end{figure*}

We demonstrated in Main Paper Fig \ref{fig:applications} how, by leveraging the advantages of our joint compositing and customization training, our model can be used for subject-driven inpainting. Additionally, Fig \ref{fig:supp_versatility} illustrates how the same model can be applied to a broad range of tasks: 

\textbf{Layout-Driven Inpainting} This task takes as input a descriptive caption, a background image, and a layout specifying an inpainting region along with object-specific bounding boxes for objects referenced in the caption. The model inpaints the selected region of the background image, ensuring alignment with the textual description while positioning objects according to the provided layout.

\textbf{Multi-Object Compositing} In addition to the inputs required for layout-driven inpainting, this task includes an image for each object corresponding to the provided bounding boxes. The model maintains the identity of these objects while enabling reposing and view synthesis, producing a cohesive composited image. 

\textbf{Layout-Driven Generation} In this case, no background image is provided. This task uses only a descriptive caption and bounding boxes specifying object positions as inputs. The model generates a full image that aligns with the caption while placing objects in the specified locations. 

\textbf{Multi-Entity Subject-Driven Generation} Similar to layout-driven generation, this task uses a text caption and bounding boxes as inputs but also includes an image for each object. The model generates a complete scene that aligns with the text, places objects in their specified locations, and preserves their unique identities.


\subsection{Multi-Object Compositing and Multi-Entity Subject-Driven Generation}

\begin{figure*}[t]
    \centering
    \includegraphics[width=\linewidth]{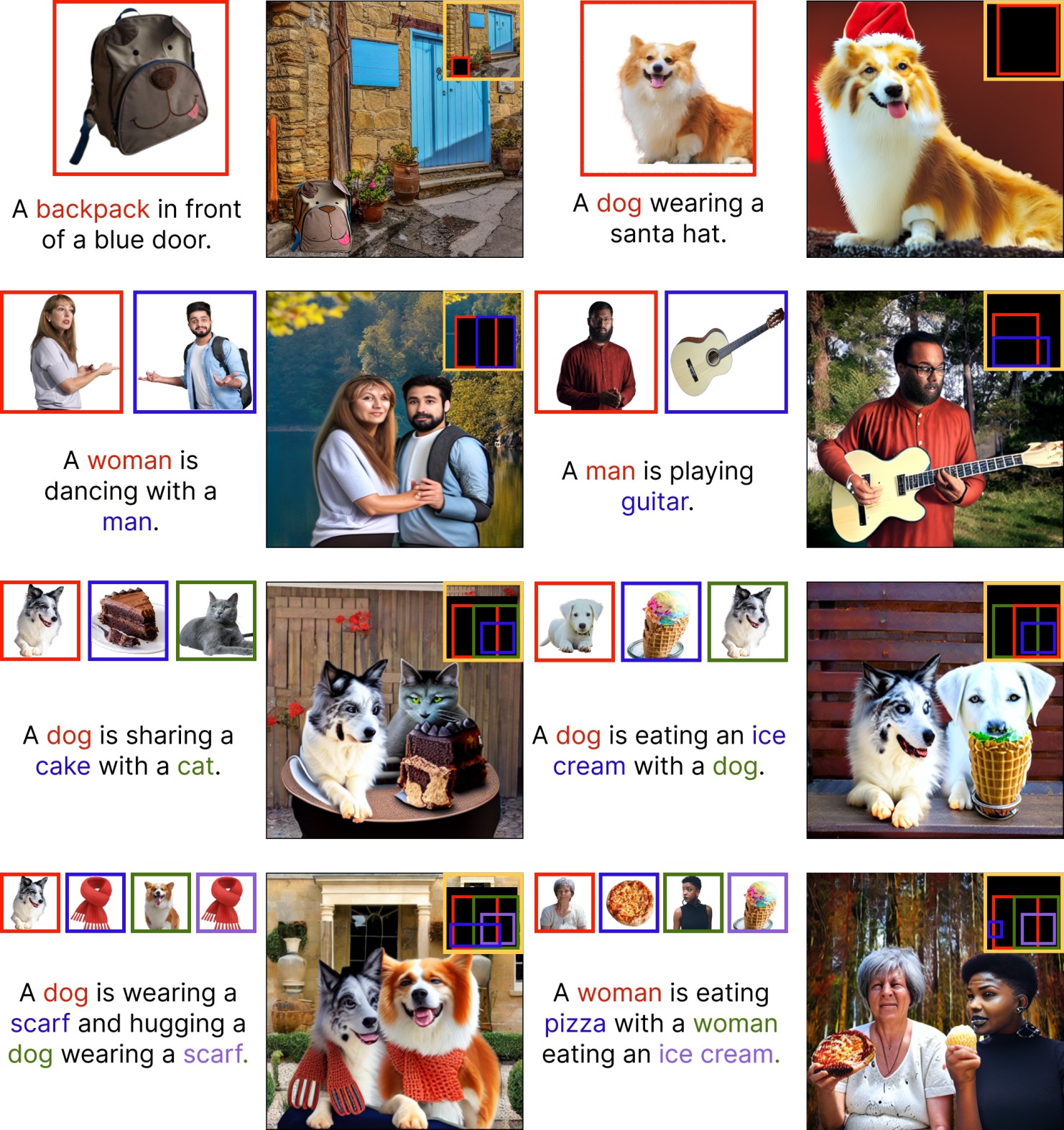}
    \caption{Visual Examples for Multi-Object Compositing (\textit{left}) and Multi-Entity Subject-Driven Generation (\textit{right}), using a variable number of grounding objects. \textit{First Row:} One Object; \textit{Second Row:} Two Objects; \textit{Third Row:} Three Objects; \textit{Forth Row:} Four Objects.}

    \label{fig:supp_customNobjs}
\end{figure*}

Fig \ref{fig:supp_customNobjs} show how the same model can be used for both multi-object compositing and multi-entity subject-driven generation, guided by a variable number of provided objects.



\end{document}